\documentclass[sigconf]{acmart}
\pdfoutput=1
\usepackage{amsmath}
\usepackage{bbm}
\usepackage{multirow}
\usepackage{makecell}
\usepackage{tabto}
\usepackage{graphicx}
\usepackage{float}
\usepackage{subfigure}
\usepackage{enumitem}

\AtBeginDocument{%
  \providecommand\BibTeX{{%
    \normalfont B\kern-0.5em{\scshape i\kern-0.25em b}\kern-0.8em\TeX}}}

\setcopyright{acmcopyright}
\copyrightyear{2018}
\acmYear{2018}
\acmDOI{10.1145/1122445.1122456}

\acmConference[Woodstock '18]{Woodstock '18: ACM Symposium on Neural
  Gaze Detection}{June 03--05, 2018}{Woodstock, NY}
\acmBooktitle{Woodstock '18: ACM Symposium on Neural Gaze Detection,
  June 03--05, 2018, Woodstock, NY}
\acmPrice{15.00}
\acmISBN{978-1-4503-XXXX-X/18/06}

\begin{document}

\title{Self-supervised Heterogeneous Graph Neural Network with Co-contrastive Learning}

\author{Xiao Wang}
\email{xiaowang@bupt.edu.cn}
\affiliation{%
  \institution{Beijing University of Posts and \\Telecommunications}
  \city{Beijing}
  \country{China}
}

\author{Nian Liu}
\email{nianliu@bupt.edu.cn}
\affiliation{%
  \institution{Beijing University of Posts and \\Telecommunications}
  \city{Beijing}
  \country{China}
}

\author{Hui Han}
\email{hanhui@bupt.edu.cn}
\affiliation{%
  \institution{Beijing University of Posts and \\Telecommunications}
  \city{Beijing}
  \country{China}
}

\author{Chuan Shi}
\authornote{Corresponding author.}
\email{shichuan@bupt.edu.cn}
\affiliation{%
  \institution{Beijing University of Posts and \\Telecommunications}
  \city{Beijing}
  \country{China}
}

\begin{abstract}
Heterogeneous graph neural networks (HGNNs) as an emerging technique have shown superior capacity of dealing with heterogeneous information network (HIN). However, most HGNNs follow a semi-supervised learning manner, which notably limits their wide use in reality since labels are usually scarce in real applications. Recently, contrastive learning, a self-supervised method, becomes one of the most exciting learning paradigms and shows great potential when there are no labels. In this paper, we study the problem of self-supervised HGNNs and propose a novel co-contrastive learning mechanism for HGNNs, named HeCo. Different from traditional contrastive learning which only focuses on contrasting positive and negative samples, HeCo employs cross-view contrastive mechanism. Specifically, two views of a HIN (network schema and meta-path views) are proposed to learn node embeddings, so as to capture both of local and high-order structures simultaneously. Then the cross-view contrastive learning, as well as a view mask mechanism, is proposed, which is able to extract the positive and negative embeddings from two views. This enables the two views to collaboratively supervise each other and finally learn high-level node embeddings. Moreover, two extensions of HeCo are designed to generate harder negative samples with high quality, which further boosts the performance of HeCo. Extensive experiments conducted on a variety of real-world networks show the superior performance of the proposed methods over the state-of-the-arts.
\end{abstract}



\maketitle

\section{Introduction}
In the real world, heterogeneous information network (HIN) or heterogeneous graph (HG) \cite{DBLP:journals/sigkdd/SunH12} is ubiquitous, due to the capacity of modeling various types of nodes and diverse interactions between them, such as bibliographic network \cite{hu2020strategies}, biomedical network \cite{DBLP:journals/nar/DavisGJSKMWWM17} and so on. Recently, heterogeneous graph neural networks (HGNNs) have achieved great success in dealing with the HIN data, because they are able to effectively combine the mechanism of message passing with complex heterogeneity, so that the complex structures and rich semantics can be well captured. So far, HGNNs have significantly promoted the development of HIN analysis towards real-world applications, e.g., recommend system \cite{DBLP:conf/kdd/FanZHSHML19} and security system \cite{DBLP:conf/kdd/FanHZYA18}.


Basically, most HGNN studies belong to the semi-supervised learning paradigm, i.e., they usually design different heterogeneous message passing mechanisms to learn node embeddings, and then the learning procedure is supervised by a part of node labels. However, the requirement that some node labels have to be known beforehand is actually frequently violated, because it is very challenging or expensive to obtain labels in some real-world environments. For example, labeling an unknown gene accurately usually needs the enormous knowledge of molecular biology, which is not easy even for veteran researchers \cite{hu2020strategies}. Recently, self-supervised learning, aiming to spontaneously find supervised signals from the data itself, becomes a promising solution for the setting without explicit labels \cite{liu2020self}. Contrastive learning, as one typical technique of self-supervised learning, has attracted considerable attentions \cite{cpc,moco,simclr,dgi,mvgrl}. By extracting positive and negative samples in data, contrastive learning aims at maximizing the similarity between positive samples while minimizing the similarity between negative samples. In this way, contrastive learning is able to learn the discriminative embeddings even without labels.
Despite the wide use of contrastive learning in computer vision \cite{simclr, moco} and natural language processing \cite{nlp1,nlp2}, little effort has been made towards investigating the great potential on HIN.

In practice, designing heterogeneous graph neural networks with contrastive learning is non-trivial, we need to carefully consider the characteristics of HIN and contrastive learning. This requires us to address the following three fundamental problems:

\textit{(1) How to design a heterogeneous contrastive mechanism.} A HIN consists of multiple types of nodes and relations, which naturally implies it possesses very complex structures. For example, meta-path, the composition of multiple relations, is usually used to capture the long-range structure in a HIN \cite{pathsim}. Different meta-paths represent different semantics, each of which reflects one aspect of HIN. To learn an effective node embedding which can fully encode these semantics, performing contrastive learning only on single meta-path view \cite{dmgi} is actually distant from sufficient. Therefore, investigating the heterogeneous cross-view contrastive mechanism is especially important for HGNNs.

\textit{(2) How to select proper views in a HIN.} As mentioned before, cross-view contrastive learning is desired for HGNNs. Despite that one can extract many different views from a HIN because of the heterogeneity, one fundamental requirement is that the selected views should cover both of the local and high-order structures. Network schema, a meta template of HIN \cite{DBLP:journals/sigkdd/SunH12}, reflects the direct connections between nodes, which naturally captures the local structure. By contrast, meta-path is widely used to extract the high-order structure. As a consequence, both of the network schema and meta-path structure views should be carefully considered.

\textit{(3) How to set a difficult contrastive task.} It is well known that a proper contrastive task will further promote to learn a more discriminative embedding \cite{simclr,amdim,cmc}. If two views are too similar, the supervised signal will be too weak to learn informative embedding. So we need to make the contrastive learning on these two views more complicated. For example, one strategy is to enhance the information diversity in two views, and the other is to generate harder negative samples of high quality. In short, designing a proper contrastive task is very crucial for HGNNs.

In this paper,  we study the problem of self-supervised learning on HIN and propose a novel heterogeneous graph neural network with co-contrastive learning (HeCo). Specifically, different from previous contrastive learning which contrasts original network and the corrupted network, we choose network schema and meta-path structure as two views to collaboratively supervise each other. In network schema view, the node embedding is learned by aggregating information from its direct neighbors, which is able to capture the local structure. In meta-path view, the node embedding is learned by passing messages along multiple meta-paths, which aims at capturing high-order structure. In this way, we design a novel contrastive mechanism, which captures complex structures in HIN. To make contrast harder, we propose a view mask mechanism that hides different parts of network schema and meta-path, respectively, which will further enhance the diversity of two views and help extract higher-level factors from these two views. Moreover, we propose two extensions of HeCo, which generate more negative samples with high quality. Finally, we modestly adapt traditional contrastive loss to the graph data, where a node has many positive samples rather than only one, different from methods \cite{simclr,moco} for CV. With the training going on, these two views are guided by each other and collaboratively optimize. The contributions of our work are summarized as follows:

\begin{itemize}
    \item To our best knowledge, this is the first attempt to study the self-supervised heterogeneous graph neural networks based on the cross-view contrastive learning. By contrastive learning based on cross-view manner, the high-level factors can be captured, enabling HGNNs to be better applied to real world applications without label supervision.
    \item We propose a novel heterogeneous graph neural network with co-contrastive learning, HeCo. HeCo innovatively employs network schema and meta-path views to collaboratively supervise each other, moreover, a view mask mechanism is designed to further enhance the contrastive performance. Additionally, two extensions of HeCo, named as HeCo\_GAN and HeCo\_MU, are proposed to generate negative samples with high quality.
    \item We conduct diverse experiments on four public datasets and the proposed HeCo outperforms the state-of-the-arts and even semi-supervised method, which demonstrates the effectiveness of HeCo from various aspects.
\end{itemize}

\section{Related Work}
In this section,we review some closely related studies, including heterogeneous graph neural network and contrastive learning.

\textbf{Heterogeneous Graph Neural Network}. Graph neural networks (GNNs) have attracted considerable attentions, where most of GNNs are proposed to homogeneous graphs, and the detailed surveys can be found in \cite{wu2021comprehensive}. Recently, some researchers focus on heterogeneous graphs. For example, HAN \cite{han} uses hierarchical attentions to depict node-level and semantic-level structures, and on this basis, MAGNN \cite{magnn} takes intermediate nodes of meta-paths into account. GTN \cite{gtn} is proposed to automatically identify useful connections. HGT \cite{hgt} is designed for Web-scale heterogeneous networks. In unsupervised setting, HetGNN \cite{hetegnn} samples a fixed size of neighbors, and fuses their features using LSTMs. NSHE \cite{nshe} focuses on network schema, and preserves pairwise and network schema proximity simultaneously. However, the above methods can not exploit supervised signals from data itself to learn general node embeddings.

\textbf{Contrastive Learning}. The approaches based on contrastive learning learn representations by contrasting positive pairs against negative pairs, and achieve great success \cite{cpc,simclr,moco,amdim}. Here we mainly focus on reviewing the graph related contrastive learning methods. Specifically, DGI \cite{dgi} builds local patches and global summary as positive pairs, and utilizes Infomax \cite{infomax} theory to contrast. Along this line, GMI \cite{gmi} is proposed to contrast between center node and its local patch from node features and topological structure. MVGRL \cite{mvgrl} employs contrast across views and experiments composition between different views. GCC \cite{gcc} focuses on pre-training with contrasting universally local structures from any two graphs. In heterogeneous domain, DMGI \cite{dmgi} conducts contrastive learning between original network and corrupted network on each single view, meta-path, and designs a consensus regularization to guide the fusion of different meta-paths. Nevertheless, there is a lack of methods contrasting across views in HIN so that the high-level factors can be captured.

\begin{figure}[t]
    \centering
    \includegraphics[width=\linewidth]{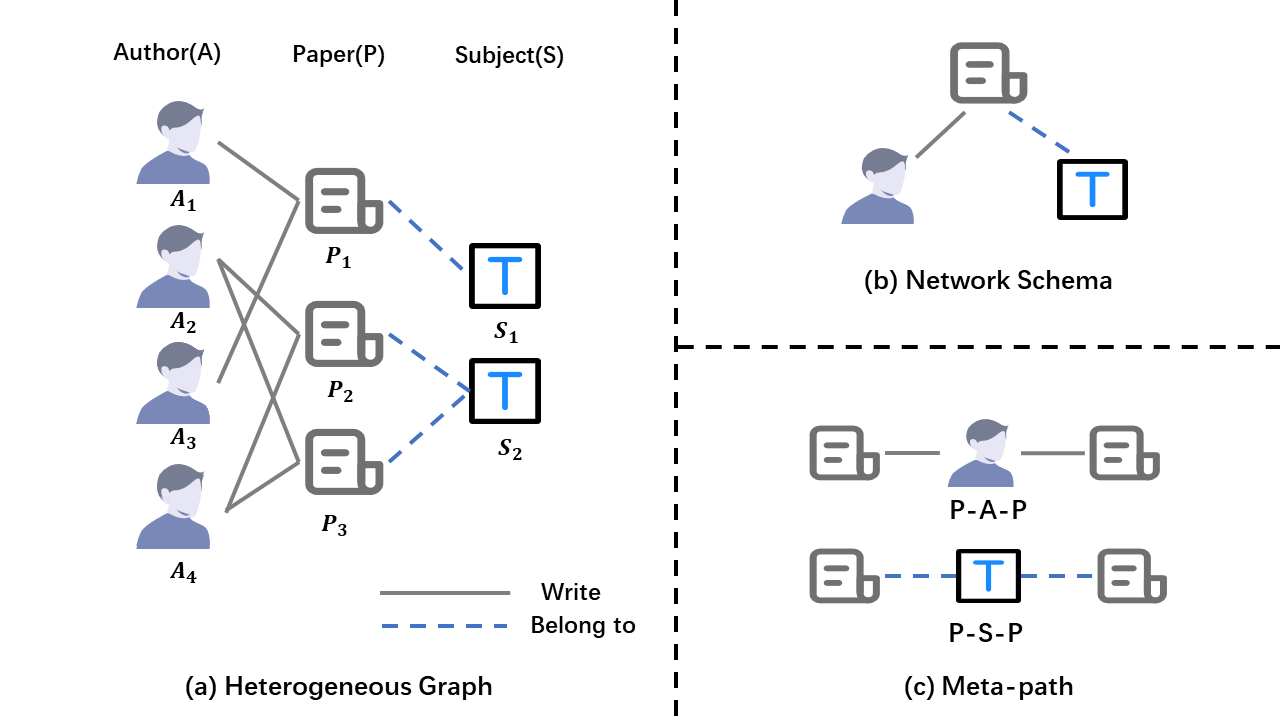}
    \caption{A toy example of HIN (ACM) and relative illustrations of meta-path and network schema.}
    \label{hg}
\end{figure}
\begin{figure*}[t]
  \centering
  \includegraphics[width=0.85\linewidth, height=70mm]{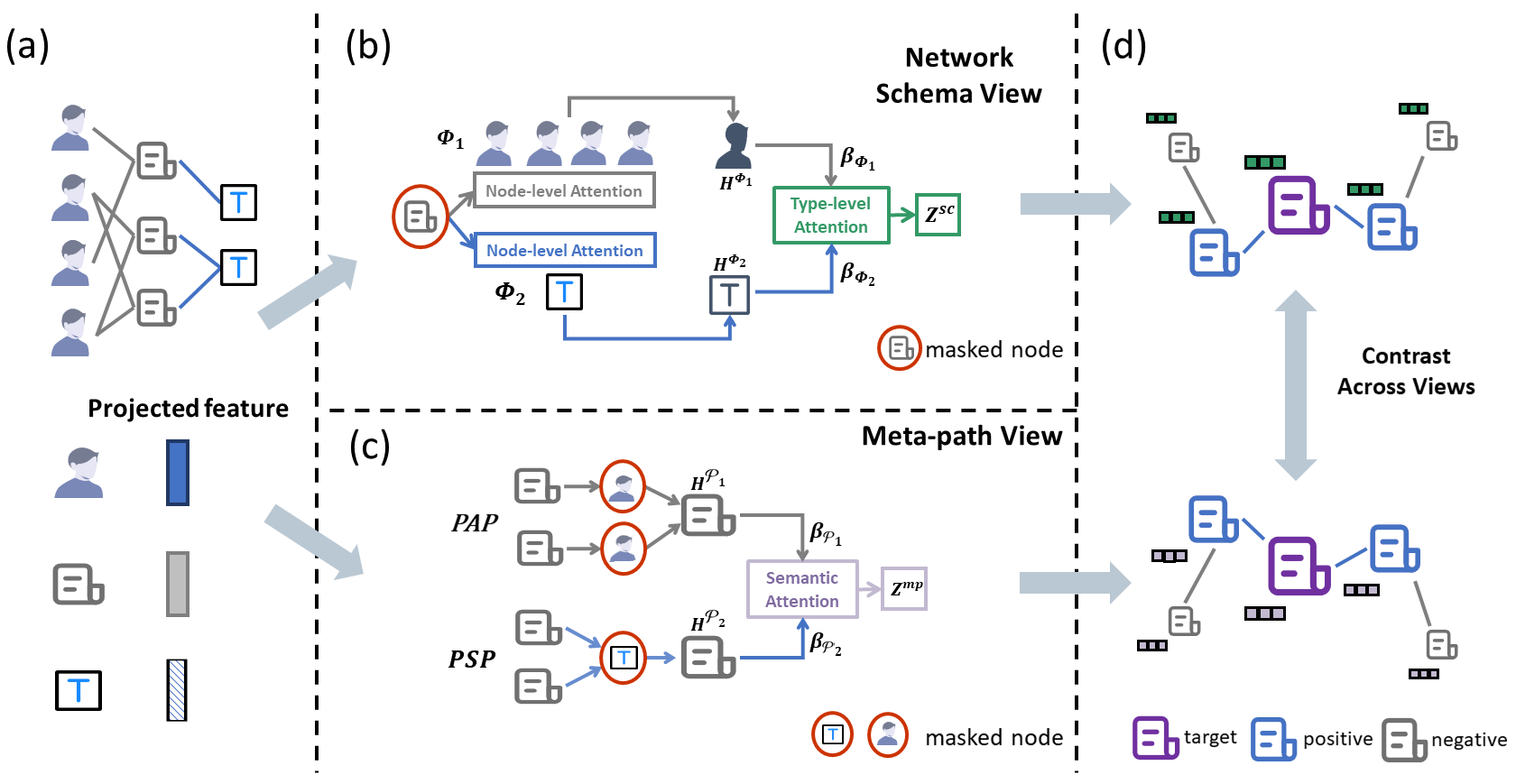}
  \caption{The overall architecture of our proposed HeCo.}
  \Description{}
  \label{model}
\end{figure*}

\section{Preliminary}
In this section, we formally define some significant concepts related to HIN as follows:

\textbf{Definition 3.1. Heterogeneous Information Network.} Heterogeneous Information Network (HIN) is defined as a network $\mathcal{G}=(\mathcal{V},\mathcal{E},\mathcal{A},\mathcal{R},\mathcal{\phi},\mathcal{\varphi})$, where $\mathcal{V}$ and $\mathcal{E}$ denote sets of nodes and edges, and it is associated with a node type mapping function $\mathcal{\phi}:\mathcal{V}\rightarrow\mathcal{A}$ and a edge type mapping function $\mathcal{\varphi}:\mathcal{E}\rightarrow\mathcal{R}$, where $\mathcal{A}$ and $\mathcal{R}$ denote sets of object and link types, and $|\mathcal{A}+\mathcal{R}|>2$.

Figure \ref{hg} (a) illustrates an example of HIN. There are three types of nodes, including author (A), paper (P) and subject (S). Meanwhile, there are two types of relations ("write" and "belong to"), i.e., author writes paper, and paper belongs to subject.

\textbf{Definition 3.2. Network Schema.} The network schema, noted as $T_G=(\mathcal{A},\mathcal{R})$, is a meta template for a HIN $\mathcal{G}$. Notice that $T_G$ is a directed graph defined over object types $\mathcal{A}$, with edges as relations from $\mathcal{R}$.

For example, Figure \ref{hg} (b) is the network schema of (a), in which we can know that paper is written by author and belongs to subject. Network schema is used to describe the direct connections between different nodes, which represents local structure.

\textbf{Definition 3.3. Meta-path.} A meta-path $\mathcal{P}$ is defined as a path, which is in the form of $A_1\stackrel{R_1}{\longrightarrow}A_2\stackrel{R_2}{\longrightarrow}\dots\stackrel{R_l}{\longrightarrow}A_{l+1}$ (abbreviated as $A_1A_2\dots A_{l+1}$), which describes a composite relation $R=R_1\circ R_2\circ \dots\circ R_l$ between node types $A_1$ and $A_{l+1}$, where $\circ$ denotes the composition operator on relations.

For example, Figure \ref{hg} (c) shows two meta-paths extracted from HIN in Figure \ref{hg} (a). PAP describes that two papers are written by the same author, and PSP describes that two papers belong to the same subject. Because meta-path is the combination of multiple relations, it contains complex semantics, which is regarded as high-order structure.

\section{The Proposed Model: H\lowercase{e}C\lowercase{o}}
In this section, we propose HeCo, a novel heterogeneous graph neural network with co-contrastive learning, and the overall architecture is shown in Figure \ref{model}. Our model encodes nodes from network schema view and meta-path view, which fully captures the structures of HIN. During the encoding, we creatively involve a view mask mechanism, which makes these two views complement and supervise mutually. With the two view-specific embeddings, we employ a contrastive learning across these two views. Given the high correlation between nodes, we redefine the positive samples of a node in HIN and design a optimization strategy specially.

\subsection{Node Feature Transformation}

Because there are different types of nodes in a HIN, their features usually lie in different spaces. So first, we need to project features of all types of nodes into a common latent vector space, as shown in Figure \ref{model} (a). Specifically, for a node $i$ with type $\phi_i$, we design a type-specific mapping matrix $W_{\phi_i}$ to transform its feature $x_i$ into common space as follows:
\begin{equation}
  h_i=\sigma\left(W_{\phi_i} \cdot x_i+b_{\phi_i}\right),
\end{equation}
where $h_i\in\mathbb{R}^{d\times1}$ is the projected feature of node $i$, $\sigma(\cdot)$ is an activation function, and $b_{\phi_i}$ denotes as vector bias, respectively.

\subsection{Network Schema View Guided Encoder}

Now we aim to learn the embedding of node $i$ under network schema view, illustrated as Figure \ref{model} (b). According to network schema, we assume that the target node $i$ connects with $S$ other types of nodes$\{\Phi_1, \Phi_2,\dots,\Phi_S\}$, so the neighbors with type $\Phi_m$ of node $i$ can be defined as $N_i^{\Phi_m}$. For node $i$, different types of neighbors contribute differently to its embedding, and so do the different nodes with the same type. So, we employ attention mechanism here in node-level and type-level to hierarchically aggregate messages from other types of neighbors to target node $i$. Specifically, we first apply node-level attention to fuse neighbors with type $\Phi_m$:
\begin{equation}
    h_i^{\Phi_m}=\sigma\left(\sum\limits_{j \in N_i^{\Phi_m}}\alpha_{i,j}^{\Phi_m} \cdot h_j\right),
    \label{sc_1}
\end{equation}
where $\sigma$ is a nonlinear activation, $h_j$ is the projected feature of node $j$, and $\alpha_{i,j}^{\Phi_m}$ denotes the attention value of node $j$ with type $\Phi_m$ to node $i$. It can be calculated as follows:
\begin{equation}
    \alpha_{i,j}^{\Phi_m}=\frac{\exp\left(LeakyReLU\left(\textbf{a}_{\Phi_m}^\top\cdot[h_i||h_j]\right)\right)}{\sum\limits_{l\in N_i^{\Phi_m}} \exp\left(LeakyReLU\left(\textbf{a}_{\Phi_m}^\top\cdot[h_i||h_l]\right)\right)},
    \label{sc_2}
\end{equation}
where $\textbf{a}_{\Phi_m}\in \mathbb{R}^{2d\times1}$ is the node-level attention vector for $\Phi_m$ and || denotes concatenate operation. Please notice that in practice, we do not aggregate the information from all the neighbors in $N_i^{\Phi_m}$, but we randomly sample a part of neighbors every epoch. Specifically, if the number of neighbors with type $\Phi_m$ exceeds a predefined threshold $T_{\Phi_m}$, we unrepeatably select $T_{\Phi_m}$ neighbors as $N_i^{\Phi_m}$, otherwise the $T_{\Phi_m}$ neighbors are selected repeatably. In this way, we ensure that every node aggregates the same amount of information from neighbors, and promote the diversity of embeddings in each epoch under this view, which will make the following contrast task more challenging.

Once we get all type embeddings $\{h_i^{\Phi_1}, ..., h_i^{\Phi_S}\}$, we utilize type-level attention to fuse them together to get the final embedding $z_i^{sc}$ for node i under network schema view. First, we measure the weight of each node type as follows:
\begin{equation}
\begin{aligned}
    w_{\Phi_m}&=\frac{1}{|V|}\sum\limits_{i\in V} \textbf{a}_{sc}^\top \cdot \tanh\left(\textbf{W}_{sc}h_i^{\Phi_m}+\textbf{b}_{sc}\right),\\
    \beta_{\Phi_m}&=\frac{\exp\left(w_{\Phi_m}\right)}{\sum_{i=1}^S\exp\left(w_{\Phi_i}\right)},
    \label{sc_4}
\end{aligned}
\end{equation}
where $V$ is the set of target nodes, \textbf{W}$_{sc}$ $\in\mathbb{R}^{d\times d}$ and \textbf{b}$_{sc}$ $\in\mathbb{R}^{d\times1}$ are learnable parameters, and $\textbf{a}_{sc}$ denotes type-level attention vector. $\beta_{\Phi_m}$ is interpreted as the importance of type $\Phi_m$ to target node $i$. So, we weighted sum the type embeddings to get $z_i^{sc}$:
\begin{equation}
    z_i^{sc}=\sum_{m=1}^S \beta_{\Phi_m}\cdot h_i^{\Phi_m}.
    \label{sc_5}
\end{equation}

\subsection{Meta-path View Guided Encoder}

Here we aim to learn the node embedding in the view of high-order meta-path structure, described in Figure \ref{model} (c). Specifically, given a meta-path $\mathcal{P}_n$ from $M$ meta-paths $\{\mathcal{P}_1,\mathcal{P}_2,\dots,\mathcal{P}_M\}$ that start from node $i$, we can get the meta-path based neighbors $N_i^{\mathcal{P}_n}$. For example, as shown in Figure \ref{hg} (a), $P_2$ is a neighbor of $P_3$ based on meta-path $PAP$. Each meta-path represents one semantic similarity, and we apply meta-path specific GCN \cite{gcn} to encode this characteristic:
\begin{equation}
    h_i^{\mathcal{P}_n}=\frac{1}{d_i+1}h_i+\sum\limits_{j\in{N_i^{\mathcal{P}_n}}}\frac{1}{\sqrt{(d_i+1)(d_j+1)}}h_j,
\end{equation}
where $d_i$ and $d_j$ are degrees of node $i$ and $j$, and $h_i$ and $h_j$ are their projected features, respectively. With $M$ meta-paths, we can get $M$ embeddings $\{h_i^{\mathcal{P}_1}, ..., h_i^{\mathcal{P}_M}\}$ for node $i$. Then we utilize semantic-level attentions to fuse them into the final embedding $z_i^{mp}$ under the meta-path view:
\begin{equation}
    z_i^{mp}=\sum_{n=1}^M \beta_{\mathcal{P}_n}\cdot h_i^{\mathcal{P}_n},
\end{equation}
where $\beta_{\mathcal{P}_n}$ weighs the importance of meta-path $\mathcal{P}_n$, which is calculated as follows:
\begin{equation}
\begin{aligned}
    w_{\mathcal{P}_n}&=\frac{1}{|V|}\sum\limits_{i\in V} \textbf{a}_{mp}^\top \cdot \tanh\left(\textbf{W}_{mp}h_i^{\mathcal{P}_n}+\textbf{b}_{mp}\right),\\
    \beta_{\mathcal{P}_n}&=\frac{\exp\left(w_{\mathcal{P}_n}\right)}{\sum_{i=1}^M\exp\left(w_{\mathcal{P}_i}\right)},
\end{aligned}
\end{equation}
where \textbf{W}$_{mp}$ $\in\mathbb{R}^{d\times d}$ and \textbf{b}$_{mp}$ $\in\mathbb{R}^{d\times1}$ are the learnable parameters, and $\textbf{a}_{mp}$ denotes the semantic-level attention vector.

\begin{figure}[t]
    \centering
    \includegraphics[width=\linewidth]{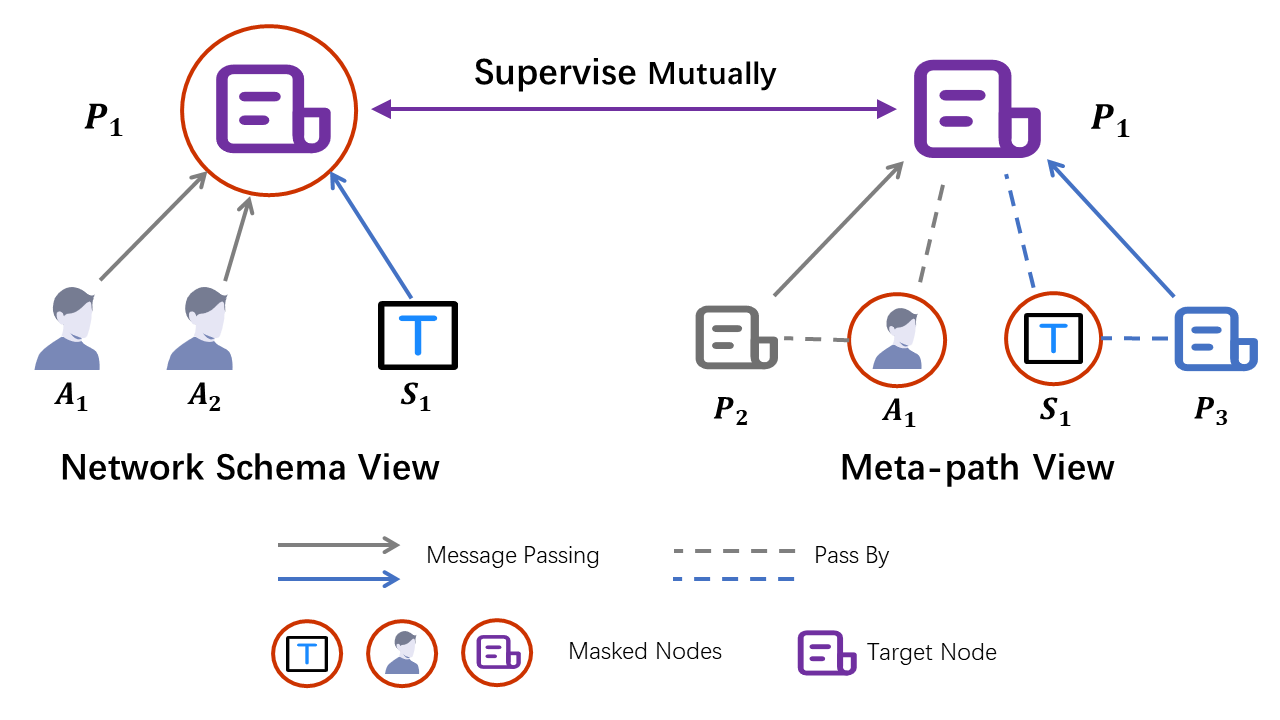}
    \caption{A schematic diagram of view mask mechanism.}
    \label{view}
\end{figure}

\subsection{View Mask Mechanism}
During the generation of $z_i^{sc}$ and $z_i^{mp}$, we design a view mask mechanism that hides different parts of network schema and meta-path views, respectively. In particular, we give a schematic diagram on ACM in Figure \ref{view}, where the target node is $P_1$. In the process of network schema encoding, $P_1$ only aggregates its neighbors including authors $A_1, A_2$ and subject $S_1$ into $z_1^{sc}$, but the message from itself is masked. While in the process of meta-path encoding, message only passes along meta-paths (e.g. PAP, PSP) from $P_2$ and $P_3$ to target $P_1$ to generate $z_1^{mp}$, while the information of intermediate nodes $A_1$ and $S_1$ are discarded. Therefore, the embeddings of node $P_1$ learned from these two parts are correlated but also complementary. They can supervise the training of each other, which presents a collaborative trend.

\subsection{Collaboratively Contrastive Optimization}
After getting the $z_i^{sc}$ and $z_i^{mp}$ for node $i$ from the above two views, we feed them into a MLP with one hidden layer to map them into the space where contrastive loss is calculated:
\begin{equation}
\begin{aligned}
    z_i^{sc}\_proj &= W^{(2)}\sigma\left(W^{(1)}z_i^{sc}+b^{(1)}\right)+b^{(2)},\\
    z_i^{mp}\_proj &= W^{(2)}\sigma\left(W^{(1)}z_i^{mp}+b^{(1)}\right)+b^{(2)},
\end{aligned}
\end{equation}
where $\sigma$ is ELU non-linear function. It should be pointed out that $\{W^{(2)},W^{(1)},b^{(2)},b^{(1)}\}$ are shared by two views embeddings. Next, when calculate contrastive loss, we need to define the positive and negative samples in a HIN.
In computer vision, generally, one image only considers its augmentations as positive samples, and treats other images as negative samples \cite{moco,amdim,cpc}. In a HIN, given a node under network schema view, we can simply define its embedding learned by meta-path view as the positive sample. However, consider that the nodes are usually highly-correlated because of edges, we propose a new positive selection strategy, i.e., if two nodes are connected by many meta-paths, they are positive samples, as shown in Figure \ref{model} (d) where links between papers represent that they are positive samples of each other. One advantage of such strategy is that the selected positive samples can well reflect the local structure of the target node.

For node $i$ and $j$, we first define a function $\mathbb{C}_i(\cdot)$ to count the number of meta-paths connecting these two nodes:
\begin{equation}
    \mathbb{C}_i(j) = \sum\limits_{n=1}^M \mathbbm{1}\left(j\in N_i^{\mathcal{P}_n}\right),
    \label{s_matrix}
\end{equation}
where $\mathbbm{1}(\cdot)$ is the indicator function. Then we construct a set $S_i=\{j|j\in V\ and\ \mathbb{C}_i(j)\neq 0\}$ and sort it in the descending order based on the value of $\mathbb{C}_i(\cdot)$. Next we set a threshold $T_{pos}$, and if $|S_i|\textgreater T_{pos}$, we select first $T_{pos}$ nodes from $S_i$ as positive samples of $i$, denotes as $\mathbb{P}_i$, otherwise all nodes in $S_i$ are retained. And we naturally treat all left nodes as negative samples of $i$, denotes as $\mathbb{N}_i$.

With the positive sample set $\mathbb{P}_i$ and negative sample set $\mathbb{N}_i$, we have the following contrastive loss under network schema view:
\begin{equation}
\label{cl}
    \mathcal{L}_i^{sc}=-\log\frac{\sum_{j\in\mathbb{P}_i}exp\left(sim\left(z_i^{sc}\_proj,z_j^{mp}\_proj\right)/\tau\right)}{\sum_{k\in\{\mathbb{P}_i\bigcup\mathbb{N}_i\}}exp\left(sim\left(z_i^{sc}\_proj,z_k^{mp}\_proj\right)/\tau\right)},
\end{equation}
where $sim(u,v)$ denotes the cosine similarity between two vectors u and v, and $\tau$ denotes a temperature parameter. We can see that different from traditional contrastive loss \cite{simclr,moco}, which usually only focuses on one positive pair in the numerator of eq.\eqref{cl}, here we consider multiple positive pairs. Also please note that for two nodes in a pair, the target embedding is from the network schema view ($z_i^{sc}\_proj$) and the embeddings of positive and negative samples are from the meta-path view ($z_k^{mp}\_proj$). In this way, we realize the cross-view self-supervision.

The contrastive loss $\mathcal{L}_i^{mp}$ is similar as $\mathcal{L}_i^{sc}$, but differently, the target embedding is from the meta-path view while the embeddings of positive and negative samples are from the network schema view. The overall objective is given as follows:
\begin{equation}
    \mathcal{J} = \frac{1}{|V|}\sum_{i\in V}\left[\lambda\cdot\mathcal{L}_i^{sc}+\left(1-\lambda\right)\cdot\mathcal{L}_i^{mp}\right],
\end{equation}
where $\lambda$ is a coefficient to balance the effect of two views. We can optimize the proposed model via back propagation and learn the embeddings of nodes. In the end, we use $z^{mp}$ to perform downstream tasks because nodes of target type explicitly participant into the generation of $z^{mp}$.

\subsection{Model Extension}
\label{extension}
It is well established that a harder negative sample is very important for contrastive learning \cite{mochi}. Therefore, to further improve the performance of HeCo, here we propose two extended models with new negative sample generation strategies.

\textbf{HeCo\_GAN} GAN-based models \cite{gan,graphgan,hegan} play a minimax game between a generator and a discriminator, and aim at forcing generator to generate fake samples, which can finally fool a well-trained discriminator. In HeCo, the negative samples are the nodes in original HIN. Here, we sample additional negatives from a continuous Gaussian distribution as in \cite{hegan}. Specifically, HeCo\_GAN is composed of three components: the proposed HeCo, a discriminator D and a generator G. We alternatively perform the following two steps and more details are provided in the Appendix \ref{hecogan}:

(1) We utilize two view-specific embeddings to train D and G alternatively. First, we train D to identify the embeddings from two views as positives and that generated from G as negatives. Then, we train G to generate samples with high quality that fool D. The two steps above are alternated for some interactions to make D and G trained.

(2) We utilize a well-trained G to generate samples, which can be viewed as the new negative samples with high quality. Then, we continue to train HeCo with the newly generated and original negative samples for some epochs.
\begin{table}
  \caption{The statistics of the datasets}
  \label{statistics}
  \setlength{\tabcolsep}{3mm}{
  \begin{tabular}{|c|c|c|c|}
    \hline
        Dataset & Node & Relation & Meta-path\\
    \hline
    ACM & \makecell*[c]{paper (P):4019\\author (A):7167\\subject (S):60} & \makecell*[c]{P-A:13407\\P-S:4019} & \makecell*[c]{PAP\\PSP} \\
    \hline
    DBLP & \makecell*[c]{author (A):4057\\paper (P):14328\\conference (C):20\\term (T):7723} & \makecell*[c]{P-A:19645\\P-C:14328\\P-T:85810} & \makecell*[c]{APA\\APCPA\\APTPA} \\
    \hline
    Freebase & \makecell*[c]{movie (M):3492\\actor (A):33401\\direct (D):2502\\writer (W):4459} & \makecell*[c]{M-A:65341\\M-D:3762\\M-W:6414} & \makecell*[c]{MAM\\MDM\\MWM} \\
    \hline
    AMiner & \makecell*[c]{paper (P):6564\\author (A):13329\\reference (R):35890} & \makecell*[c]{P-A:18007\\P-R:58831} & \makecell*[c]{PAP\\PRP} \\
    \hline
\end{tabular}}
\end{table}

\textbf{HeCo\_MU} MixUp \cite{mixup} is proposed to efficiently improve results in supervised learning by adding arbitrary two samples to create a new one. MoCHi \cite{mochi} introduces this strategy into contrastive learning , who mixes the hard negatives to make harder negatives. Inspired by them, we bring this strategy into HIN field for the first time. We can get cosine similarities between node $i$ and nodes from $\mathbb{N}_i$ during calculating eq.\eqref{cl}, and sort them in the descending order. Then, we select first top k negative samples as the hardest negatives, and randomly add them to create new k negatives, which are involved in training. It is worth mentioning that there are no learnable parameters in this version, which is very efficient.

\begin{table*}[t]
  \caption{Quantitative results (\%$\pm\sigma$) on node classification.}
  \label{fenlei}
  \resizebox{\textwidth}{!}{
  \begin{tabular}{c|c|c|cccccccc|c}
    \hline
    Datasets & Metric & Split & GraphSAGE & GAE & Mp2vec & HERec & HetGNN & HAN & DGI & DMGI & HeCo\\
    \hline
    \multirow{9}{*}{ACM}&
    \multirow{3}{*}{Ma-F1}
    &20&47.13$\pm$4.7&62.72$\pm$3.1&51.91$\pm$0.9&55.13$\pm$1.5&72.11$\pm$0.9&85.66$\pm$2.1&79.27$\pm$3.8&87.86$\pm$0.2&\textbf{88.56$\pm$0.8}\\
    &&40&55.96$\pm$6.8&61.61$\pm$3.2&62.41$\pm$0.6&61.21$\pm$0.8&72.02$\pm$0.4&87.47$\pm$1.1&80.23$\pm$3.3&86.23$\pm$0.8&\textbf{87.61$\pm$0.5}\\
    &&60&56.59$\pm$5.7&61.67$\pm$2.9&61.13$\pm$0.4&64.35$\pm$0.8&74.33$\pm$0.6&88.41$\pm$1.1&80.03$\pm$3.3&87.97$\pm$0.4&\textbf{89.04$\pm$0.5}\\
    \cline{2-12}
    &\multirow{3}{*}{Mi-F1}
    &20&49.72$\pm$5.5&68.02$\pm$1.9&53.13$\pm$0.9&57.47$\pm$1.5&71.89$\pm$1.1&85.11$\pm$2.2&79.63$\pm$3.5&87.60$\pm$0.8&\textbf{88.13$\pm$0.8}\\
    &&40&60.98$\pm$3.5&66.38$\pm$1.9&64.43$\pm$0.6&62.62$\pm$0.9&74.46$\pm$0.8&87.21$\pm$1.2&80.41$\pm$3.0&86.02$\pm$0.9&\textbf{87.45$\pm$0.5}\\
    &&60&60.72$\pm$4.3&65.71$\pm$2.2&62.72$\pm$0.3&65.15$\pm$0.9&76.08$\pm$0.7&88.10$\pm$1.2&80.15$\pm$3.2&87.82$\pm$0.5&\textbf{88.71$\pm$0.5}\\
    \cline{2-12}
    &\multirow{3}{*}{AUC}
    &20&65.88$\pm$3.7&79.50$\pm$2.4&71.66$\pm$0.7&75.44$\pm$1.3&84.36$\pm$1.0&93.47$\pm$1.5&91.47$\pm$2.3&\textbf{96.72$\pm$0.3}&96.49$\pm$0.3\\
    &&40&71.06$\pm$5.2&79.14$\pm$2.5&80.48$\pm$0.4&79.84$\pm$0.5&85.01$\pm$0.6&94.84$\pm$0.9&91.52$\pm$2.3&96.35$\pm$0.3&\textbf{96.40$\pm$0.4}\\
    &&60&70.45$\pm$6.2&77.90$\pm$2.8&79.33$\pm$0.4&81.64$\pm$0.7&87.64$\pm$0.7&94.68$\pm$1.4&91.41$\pm$1.9&\textbf{96.79$\pm$0.2}&96.55$\pm$0.3\\
    \hline
    \multirow{9}{*}{DBLP}&
    \multirow{3}{*}{Ma-F1}
    &20&71.97$\pm$8.4&90.90$\pm$0.1&88.98$\pm$0.2&89.57$\pm$0.4&89.51$\pm$1.1&89.31$\pm$0.9&87.93$\pm$2.4&89.94$\pm$0.4&\textbf{91.28$\pm$0.2}\\
    &&40&73.69$\pm$8.4&89.60$\pm$0.3&88.68$\pm$0.2&89.73$\pm$0.4&88.61$\pm$0.8&88.87$\pm$1.0&88.62$\pm$0.6&89.25$\pm$0.4&\textbf{90.34$\pm$0.3}\\
    &&60&73.86$\pm$8.1&90.08$\pm$0.2&90.25$\pm$0.1&90.18$\pm$0.3&89.56$\pm$0.5&89.20$\pm$0.8&89.19$\pm$0.9&89.46$\pm$0.6&\textbf{90.64$\pm$0.3}\\
    \cline{2-12}
    &\multirow{3}{*}{Mi-F1}
    &20&71.44$\pm$8.7&91.55$\pm$0.1&89.67$\pm$0.1&90.24$\pm$0.4&90.11$\pm$1.0&90.16$\pm$0.9&88.72$\pm$2.6&90.78$\pm$0.3&\textbf{91.97$\pm$0.2}\\
    &&40&73.61$\pm$8.6&90.00$\pm$0.3&89.14$\pm$0.2&90.15$\pm$0.4&89.03$\pm$0.7&89.47$\pm$0.9&89.22$\pm$0.5&89.92$\pm$0.4&\textbf{90.76$\pm$0.3}\\
    &&60&74.05$\pm$8.3&90.95$\pm$0.2&91.17$\pm$0.1&91.01$\pm$0.3&90.43$\pm$0.6&90.34$\pm$0.8&90.35$\pm$0.8&90.66$\pm$0.5&\textbf{91.59$\pm$0.2}\\
    \cline{2-12}
    &\multirow{3}{*}{AUC}
    &20&90.59$\pm$4.3&98.15$\pm$0.1&97.69$\pm$0.0&98.21$\pm$0.2&97.96$\pm$0.4&98.07$\pm$0.6&96.99$\pm$1.4&97.75$\pm$0.3&\textbf{98.32$\pm$0.1}\\
    &&40&91.42$\pm$4.0&97.85$\pm$0.1&97.08$\pm$0.0&97.93$\pm$0.1&97.70$\pm$0.3&97.48$\pm$0.6&97.12$\pm$0.4&97.23$\pm$0.2&\textbf{98.06$\pm$0.1}\\
    &&60&91.73$\pm$3.8&98.37$\pm$0.1&98.00$\pm$0.0&98.49$\pm$0.1&97.97$\pm$0.2&97.96$\pm$0.5&97.76$\pm$0.5&97.72$\pm$0.4&\textbf{98.59$\pm$0.1}\\
    \hline
    \multirow{9}{*}{Freebase}&
    \multirow{3}{*}{Ma-F1}
    &20&45.14$\pm$4.5&53.81$\pm$0.6&53.96$\pm$0.7&55.78$\pm$0.5&52.72$\pm$1.0&53.16$\pm$2.8&54.90$\pm$0.7&55.79$\pm$0.9&\textbf{59.23$\pm$0.7}\\
    &&40&44.88$\pm$4.1&52.44$\pm$2.3&57.80$\pm$1.1&59.28$\pm$0.6&48.57$\pm$0.5&59.63$\pm$2.3&53.40$\pm$1.4&49.88$\pm$1.9&\textbf{61.19$\pm$0.6}\\
    &&60&45.16$\pm$3.1&50.65$\pm$0.4&55.94$\pm$0.7&56.50$\pm$0.4&52.37$\pm$0.8&56.77$\pm$1.7&53.81$\pm$1.1&52.10$\pm$0.7&\textbf{60.13$\pm$1.3}\\
    \cline{2-12}
    &\multirow{3}{*}{Mi-F1}
    &20&54.83$\pm$3.0&55.20$\pm$0.7&56.23$\pm$0.8&57.92$\pm$0.5&56.85$\pm$0.9&57.24$\pm$3.2&58.16$\pm$0.9&58.26$\pm$0.9&\textbf{61.72$\pm$0.6}\\
    &&40&57.08$\pm$3.2&56.05$\pm$2.0&61.01$\pm$1.3&62.71$\pm$0.7&53.96$\pm$1.1&63.74$\pm$2.7&57.82$\pm$0.8&54.28$\pm$1.6&\textbf{64.03$\pm$0.7}\\
    &&60&55.92$\pm$3.2&53.85$\pm$0.4&58.74$\pm$0.8&58.57$\pm$0.5&56.84$\pm$0.7&61.06$\pm$2.0&57.96$\pm$0.7&56.69$\pm$1.2&\textbf{63.61$\pm$1.6}\\
    \cline{2-12}
    &\multirow{3}{*}{AUC}
    &20&67.63$\pm$5.0&73.03$\pm$0.7&71.78$\pm$0.7&73.89$\pm$0.4&70.84$\pm$0.7&73.26$\pm$2.1&72.80$\pm$0.6&73.19$\pm$1.2&\textbf{76.22$\pm$0.8}\\
    &&40&66.42$\pm$4.7&74.05$\pm$0.9&75.51$\pm$0.8&76.08$\pm$0.4&69.48$\pm$0.2&77.74$\pm$1.2&72.97$\pm$1.1&70.77$\pm$1.6&\textbf{78.44$\pm$0.5}\\
    &&60&66.78$\pm$3.5&71.75$\pm$0.4&74.78$\pm$0.4&74.89$\pm$0.4&71.01$\pm$0.5&75.69$\pm$1.5&73.32$\pm$0.9&73.17$\pm$1.4&\textbf{78.04$\pm$0.4}\\
    \hline
    \multirow{9}{*}{AMiner}&
    \multirow{3}{*}{Ma-F1}
    &20&42.46$\pm$2.5&60.22$\pm$2.0&54.78$\pm$0.5&58.32$\pm$1.1&50.06$\pm$0.9&56.07$\pm$3.2&51.61$\pm$3.2&59.50$\pm$2.1&\textbf{71.38$\pm$1.1}\\
    &&40&45.77$\pm$1.5&65.66$\pm$1.5&64.77$\pm$0.5&64.50$\pm$0.7&58.97$\pm$0.9&63.85$\pm$1.5&54.72$\pm$2.6&61.92$\pm$2.1&\textbf{73.75$\pm$0.5}\\
    &&60&44.91$\pm$2.0&63.74$\pm$1.6&60.65$\pm$0.3&65.53$\pm$0.7&57.34$\pm$1.4&62.02$\pm$1.2&55.45$\pm$2.4&61.15$\pm$2.5&\textbf{75.80$\pm$1.8}\\
    \cline{2-12}
    &\multirow{3}{*}{Mi-F1}
    &20&49.68$\pm$3.1&65.78$\pm$2.9&60.82$\pm$0.4&63.64$\pm$1.1&61.49$\pm$2.5&68.86$\pm$4.6&62.39$\pm$3.9&63.93$\pm$3.3&\textbf{78.81$\pm$1.3}\\
    &&40&52.10$\pm$2.2&71.34$\pm$1.8&69.66$\pm$0.6&71.57$\pm$0.7&68.47$\pm$2.2&76.89$\pm$1.6&63.87$\pm$2.9&63.60$\pm$2.5&\textbf{80.53$\pm$0.7}\\
    &&60&51.36$\pm$2.2&67.70$\pm$1.9&63.92$\pm$0.5&69.76$\pm$0.8&65.61$\pm$2.2&74.73$\pm$1.4&63.10$\pm$3.0&62.51$\pm$2.6&\textbf{82.46$\pm$1.4}\\
    \cline{2-12}
    &\multirow{3}{*}{AUC}
    &20&70.86$\pm$2.5&85.39$\pm$1.0&81.22$\pm$0.3&83.35$\pm$0.5&77.96$\pm$1.4&78.92$\pm$2.3&75.89$\pm$2.2&85.34$\pm$0.9&\textbf{90.82$\pm$0.6}\\
    &&40&74.44$\pm$1.3&88.29$\pm$1.0&88.82$\pm$0.2&88.70$\pm$0.4&83.14$\pm$1.6&80.72$\pm$2.1&77.86$\pm$2.1&88.02$\pm$1.3&\textbf{92.11$\pm$0.6}\\
    &&60&74.16$\pm$1.3&86.92$\pm$0.8&85.57$\pm$0.2&87.74$\pm$0.5&84.77$\pm$0.9&80.39$\pm$1.5&77.21$\pm$1.4&86.20$\pm$1.7&\textbf{92.40$\pm$0.7}\\
    \hline
  \end{tabular}}
\end{table*}
\section{EXPERIMENTS}
\subsection{Experimental Setup}

\textbf{Datasets}\quad We employ the following four real HIN datasets, where the basic information are summarized in Table \ref{statistics}.
\begin{itemize}
\item \textbf{ACM} \cite{nshe}. The target nodes are papers, which are divided into three classes. For each paper, there are 3.33 authors averagely, and one subject.
\item \textbf{DBLP} \cite{magnn}. The target nodes are authors, which are divided into four classes. For each author, there are 4.84 papers averagely.
\item \textbf{Freebase} \cite{freebase}. The target nodes are movies, which are divided into three classes. For each movie, there are 18.7 actors, 1.07 directors and 1.83 writers averagely.
\item \textbf{AMiner} \cite{hegan}. The target nodes are papers. We extract a subset of original dataset, where papers are divided into four classes. For each paper, there are 2.74 authors and 8.96 references averagely.
\end{itemize}
\noindent\textbf{Baselines}\quad We compare the proposed HeCo with three categories of baselines: unsupervised homogeneous methods \{ GraphSAGE \cite{GraphSAGE}, GAE \cite{gae}, DGI \cite{dgi} \}, unsupervised heterogeneous methods \{ Mp2vec \cite{mp2vec}, HERec \cite{herec}, HetGNN \cite{hetegnn}, DMGI \cite{dmgi} \}, and a semi-supervised heterogeneous method HAN \cite{han}.

\begin{figure*}
\centering
\subfigure[Mp2vec]{
\label{Mp2vec_acm}
\includegraphics[width=0.2\textwidth]{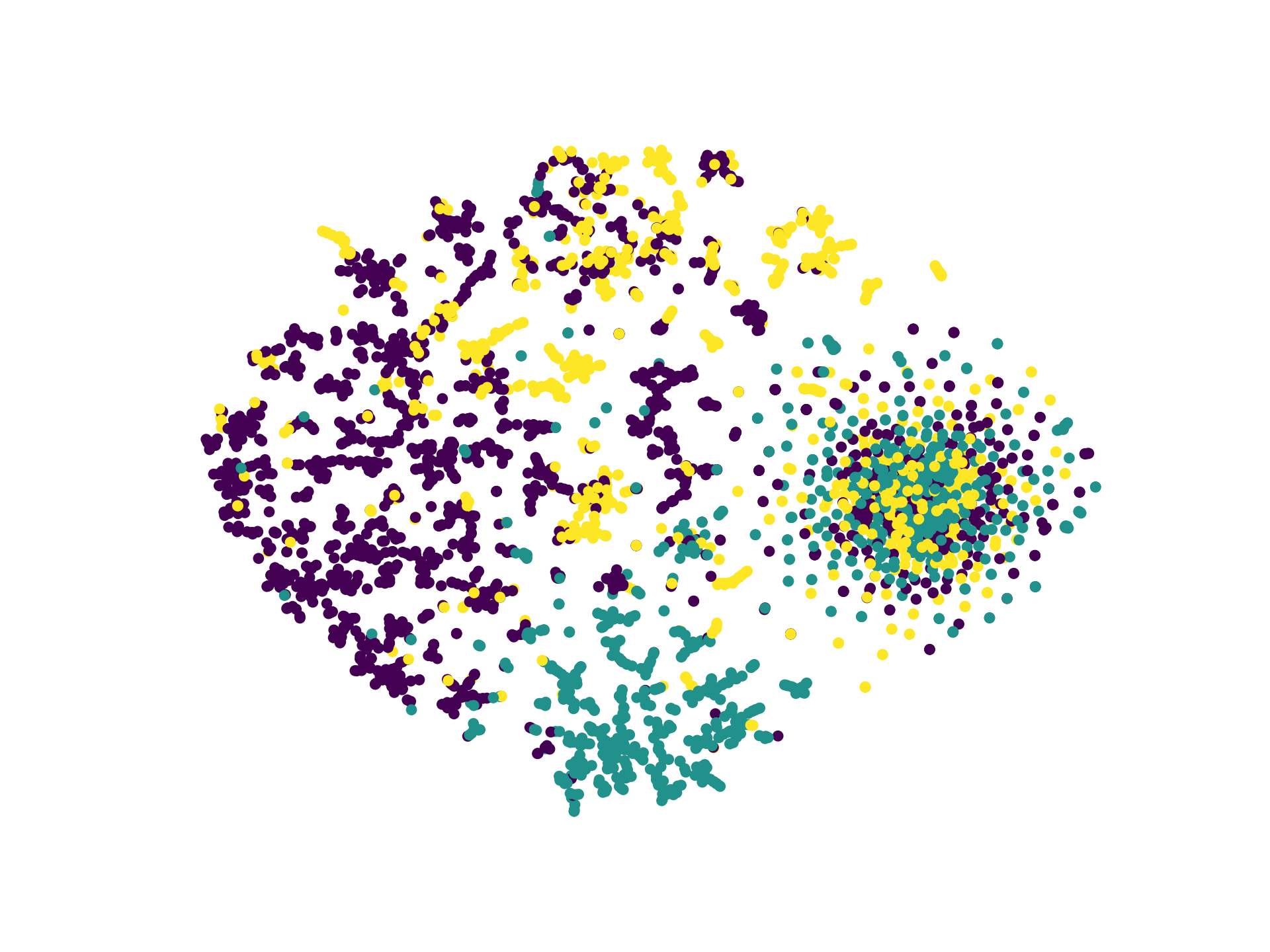}}
\subfigure[DGI]{
\label{DGI_acm}
\includegraphics[width=0.2\textwidth]{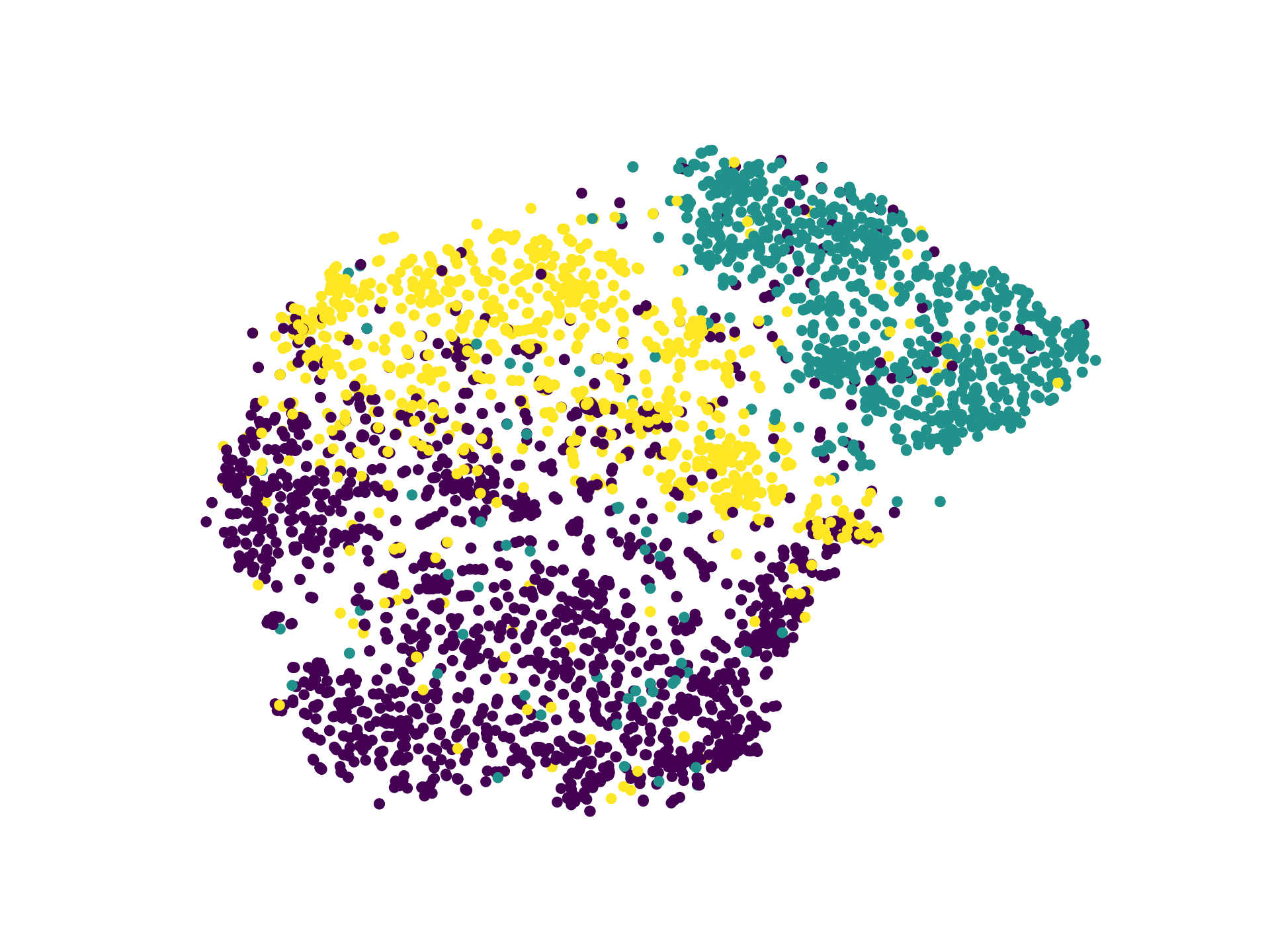}}
\subfigure[DMGI]{
\label{DMGI_v_acm}
\includegraphics[width=0.2\textwidth]{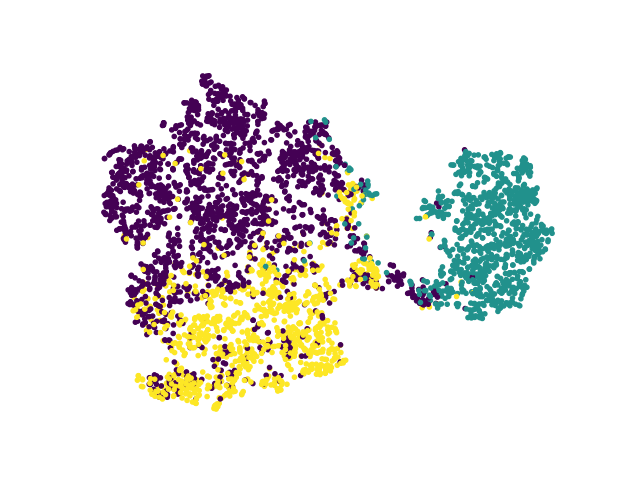}}
\subfigure[HeCo]{
\label{HeCo_v_acm}
\includegraphics[width=0.2\textwidth]{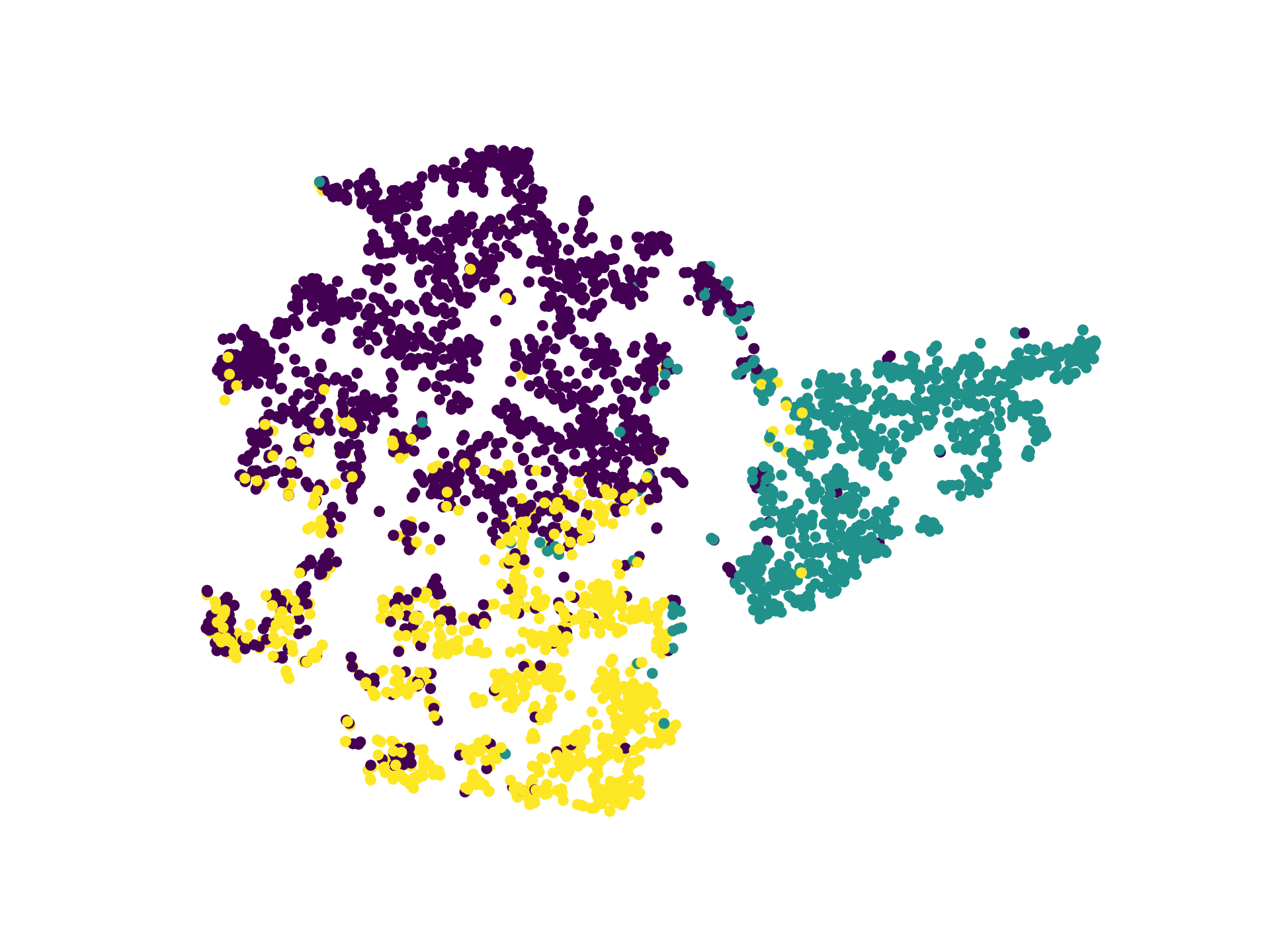}}
\caption{Visualization of the learned node embedding on ACM. The Silhouette scores for (a) (b) (c) (d) are 0.0292, 0.1862, 0.3015 and 0.3642, respectively.}
\label{visiulization}
\end{figure*}

\noindent\textbf{Implementation Details}\quad For random walk-based methods (i.e., Mp2vec, HERec, HetGNN), we set the number of walks per node to 40, the walk length to 100 and the window size to 5. For GraphSAGE, GAE, Mp2vec, HERec and DGI, we test all the meta-paths for them and report the best performance. In terms of other parameters, we follow the settings in their original papers.

For the proposed HeCo, we initialize parameters using Glorot initialization \cite{glorot2010understanding}, and train the model with Adam \cite{adam}. We search on learning rate from 1e-4 to 5e-3, and tune the patience for early stopping from 5 to 50, i.e, we stop training if the contrastive loss does not decrease for patience consecutive epochs. For dropout used on attentions in two views, we test ranging from 0.1 to 0.5 with step 0.05, and $\tau$ is tuned from 0.5 to 0.9 also with step 0.05. Moreover, we only perform aggregation one time in each view. That is to say, for meta-path view we use one-layer GCN for every meta-path, and for network schema view we only consider interactions between nodes of target type and their one-hop neighbors of other types. The source code and datasets are publicly available on Github $\footnote{\url{https://github.com/liun-online/HeCo}}$.

For all methods, we set the embedding dimension as 64 and randomly run 10 times and report the average results. For every dataset, we only use original attributes of target nodes, and assign one-hot id vectors to nodes of other types, if they are needed. For the reproducibility, we provide the specific parameter values in the supplement (Section \ref{detail}).

\subsection{Node Classification}
The learned embeddings of nodes are used to train a linear classifier. To more comprehensively evaluate our model, we choose 20, 40, 60 labeled nodes per class as training set, and select 1000 nodes as validation and 1000 as the test set respectively, for each dataset. We follow DMGI that report the test performance when the performance on validation gives the best result. We use common evaluation metrics, including Macro-F1, Micro-F1 and AUC. The results are reported in Table \ref{fenlei}. As can be seen, the proposed HeCo generally outperforms all the other methods on all datasets and all splits, even compared with HAN, a semi-supervised method.
We can also see that HeCo outperforms DMGI in most cases, while DMGI is even worse than other baselines with some settings, indicating that single-view is noisy and incomplete. So, performing contrastive learning across views is effective.
Moreover, even HAN utilizes the label information, still, HeCo performs better than it in all cases. This well indicates the great potential of cross-view contrastive learning.

\begin{table}
  \caption{Quantitative results (\%$\pm\sigma$) on node clustering.}
  \label{julei}
  \resizebox{0.47\textwidth}{!}{
  \begin{tabular}{c|cc|cc|cc|cc}
    \hline
    Datasets & \multicolumn{2}{c|}{ACM} & \multicolumn{2}{c|}{DBLP} & \multicolumn{2}{c|}{Freebase} & \multicolumn{2}{c}{AMiner} \\
    \hline
    Metrics & NMI & ARI & NMI & ARI & NMI & ARI & NMI & ARI\\
    \hline
    GraphSage&29.20&27.72&51.50&36.40&9.05&10.49&15.74&10.10\\
    GAE&27.42&24.49&72.59&77.31&19.03&14.10&28.58&20.90\\
    Mp2vec&48.43&34.65&73.55&77.70&16.47&17.32&30.80&25.26\\
    HERec&47.54&35.67&70.21&73.99&19.76&19.36&27.82&20.16\\
    HetGNN&41.53&34.81&69.79&75.34&12.25&15.01&21.46&26.60\\
    DGI&51.73&41.16&59.23&61.85&18.34&11.29&22.06&15.93\\
    DMGI&51.66&46.64&70.06&75.46&16.98&16.91&19.24&20.09\\
    \hline
    HeCo&\textbf{56.87}&\textbf{56.94}&\textbf{74.51}&\textbf{80.17}&\textbf{20.38}&\textbf{20.98}&\textbf{32.26}&\textbf{28.64}\\
    \hline
  \end{tabular}}
\end{table}

\subsection{Node Clustering}
In this task, we utilize K-means algorithm to the learned embeddings of all nodes and adopt normalized mutual information (NMI) and adjusted rand index (ARI) to assess the quality of the clustering results. To alleviate the instability due to different initial values, we repeat the process for 10 times, and report average results, shown in Table \ref{julei}. Notice that, we do not compare with HAN, because it has known the labels of training set and been guided by validation.
As we can see, HeCo consistently achieves the best results on all datasets, which proves the effectiveness of HeCo from another angle. Especially, HeCo gains about 10\% improvements on NMI and 20\% improvements on ARI on ACM, demonstrating the superiority of our model. Moreover, HeCo outperforms DMGI in all cases, further suggesting the importance of contrasting across views.

\subsection{Visualization}
To provide a more intuitive evaluation, we conduct embedding visualization on ACM dataset. We plot learnt embeddings of Mp2vec, DGI, DMGI and HeCo using t-SNE, and the results are shown in Figure \ref{visiulization}, in which different colors mean different labels.

We can see that Mp2vec and DGI present blurred boundaries between different types of nodes, because they cannot fuse all kinds of semantics. For DMGI, nodes are still mixed to some degree. The proposed HeCo correctly separates different nodes with relatively clear boundaries. Moreover, we calculate the silhouette score of different clusters, and HeCo also outperforms other three methods, demonstrating the effectiveness of HeCo again.

\subsection{Variant Analysis}
\begin{figure}[t]
\centering
    \subfigure[ACM]{
        \label{xiao_acm40}
        \centering
        \includegraphics[scale=0.28]{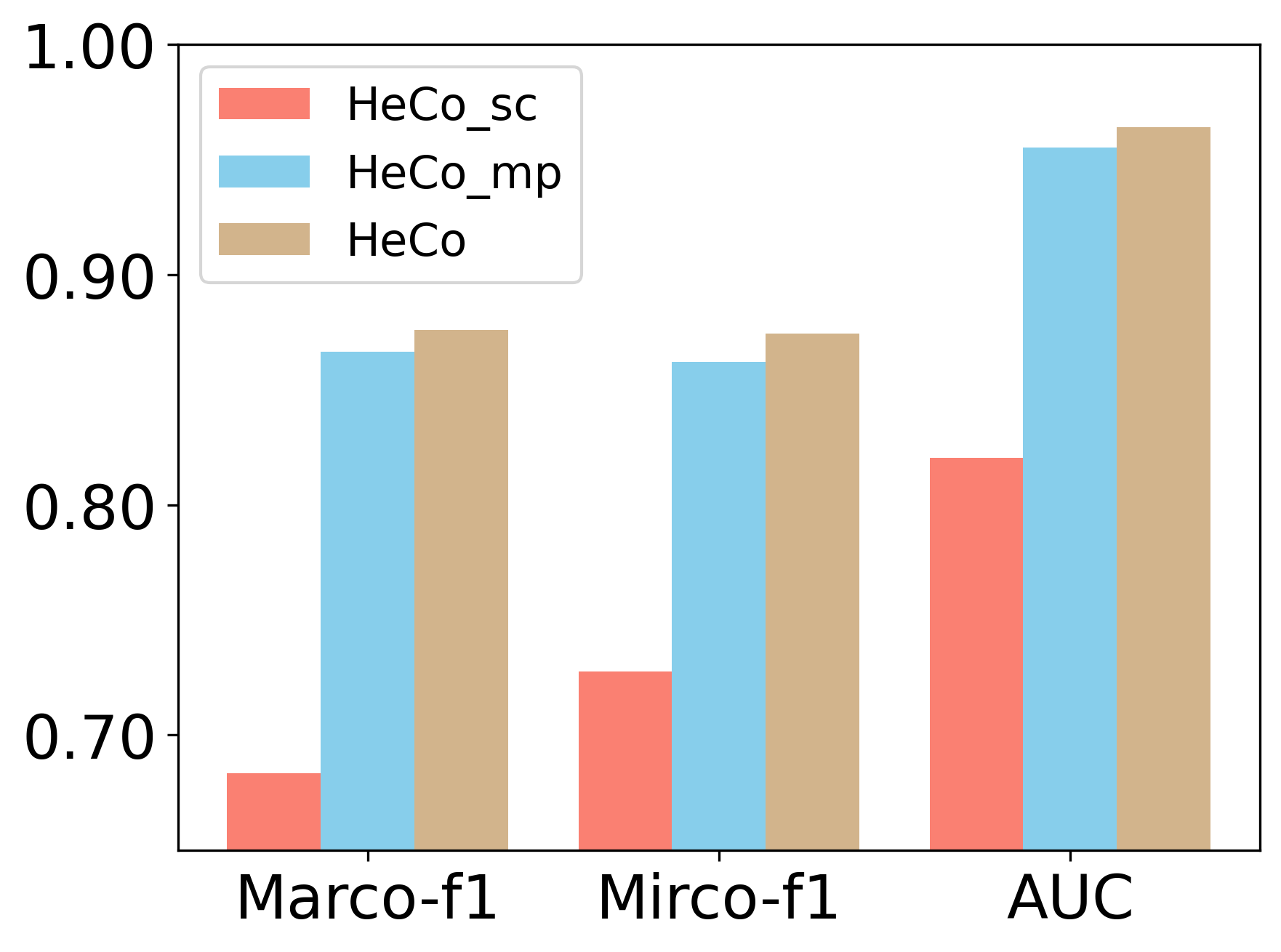}
        }
    \subfigure[DBLP]{
        \label{xiao_dblp40}
        \centering
        \includegraphics[scale=0.28]{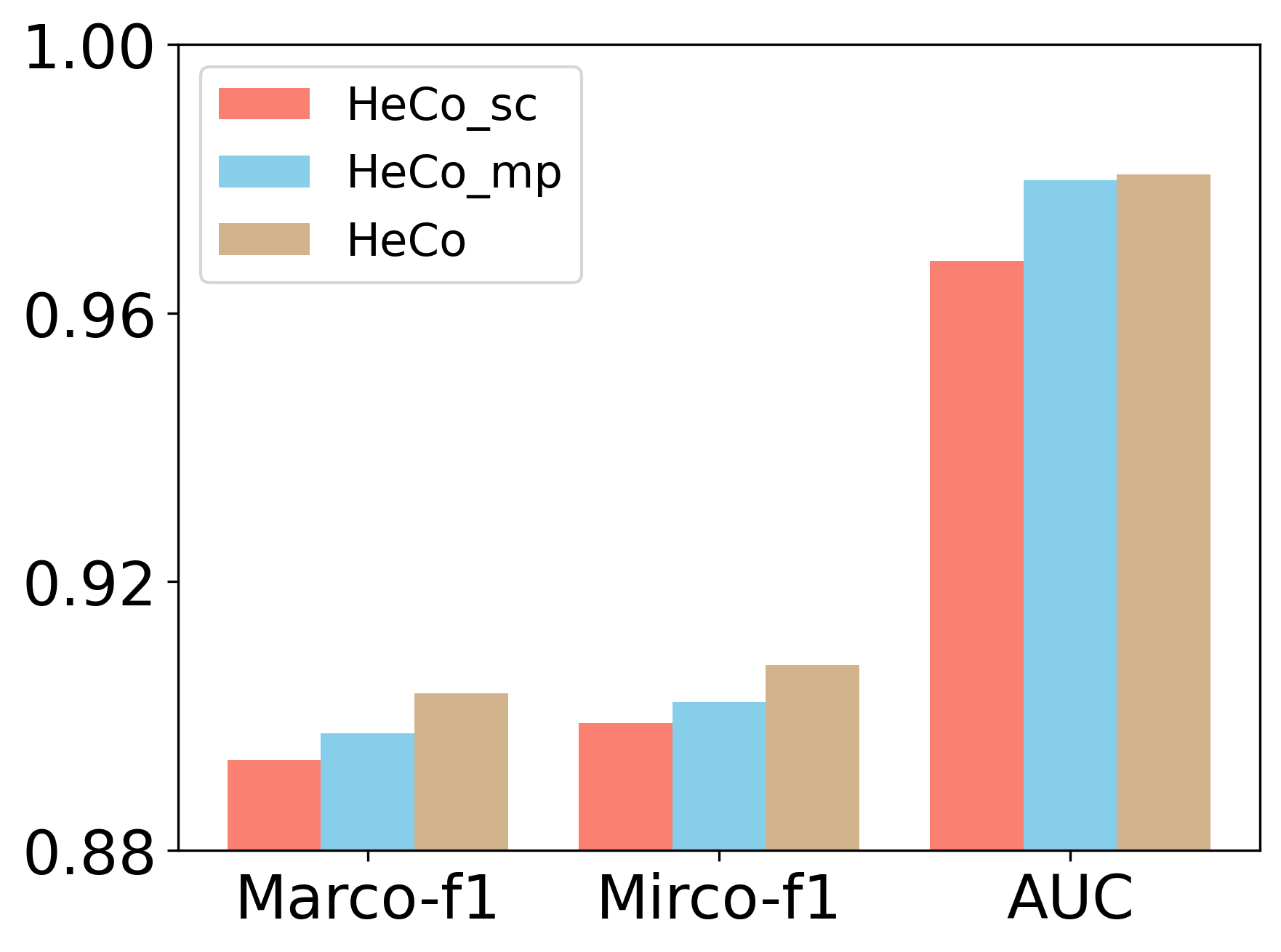}
        }
\caption{The comparison of HeCo and its variants.}
\label{variants}
\end{figure}

In this section, we design two variants of proposed HeCo: HeCo\_sc and HeCo\_mp. For variant HeCo\_sc, nodes are only encoded in network schema view, and the embeddings of corresponding positive and negatives samples also come from network schema view, rather than meta-path view. For variant HeCo\_mp, the practice is similar, where we only focus on meta-path view and neglect network schema view. We conduct comparison between them and HeCo on ACM and DBLP, and report the results of 40 labeled nodes per class, which are given in Figure \ref{variants}.

From Figure \ref{variants}, some conclusions are got as follows: (1) The results of HeCo are consistently better than two variants, indicating the effectiveness and necessity of the cross-view contrastive learning. (2) The performance of HeCo\_mp is also very competitive, especially in DBLP dataset, which demonstrates that meta-path is a powerful tool to handle the heterogeneity due to capacity of capturing semantic information between nodes. (3) HeCo\_sc performs worse than other methods, which makes us realize the necessity of involving the features of target nodes into embeddings if contrast is done only in a single view.

\subsection{Collaborative Trend Analysis}
\begin{figure}[h]
\centering
\subfigure[ACM]{
\label{xuan_acm}
\centering
\includegraphics[scale=0.23]{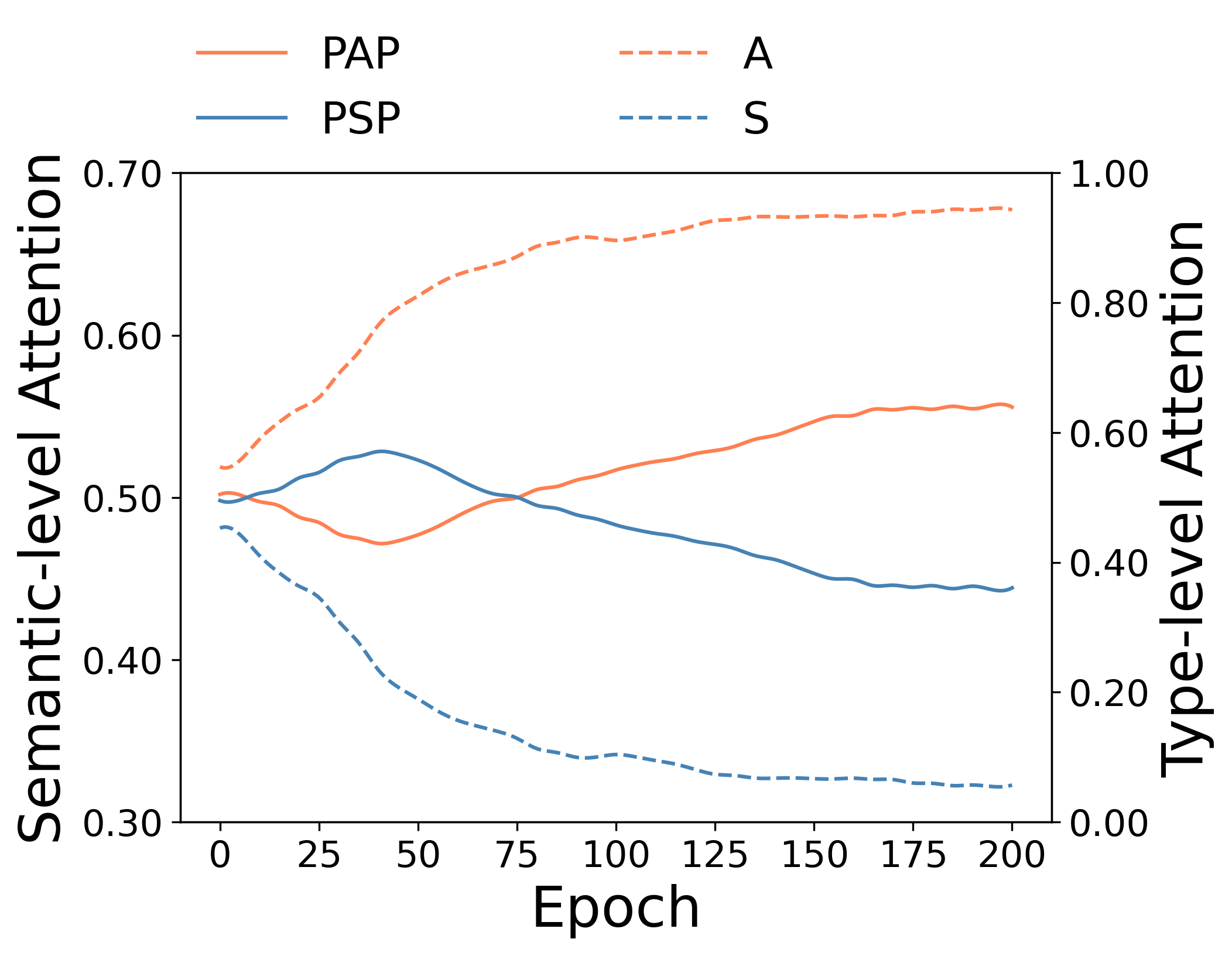}}
\subfigure[AMiner]{
\label{xuan_aminer}
\centering
\includegraphics[scale=0.23]{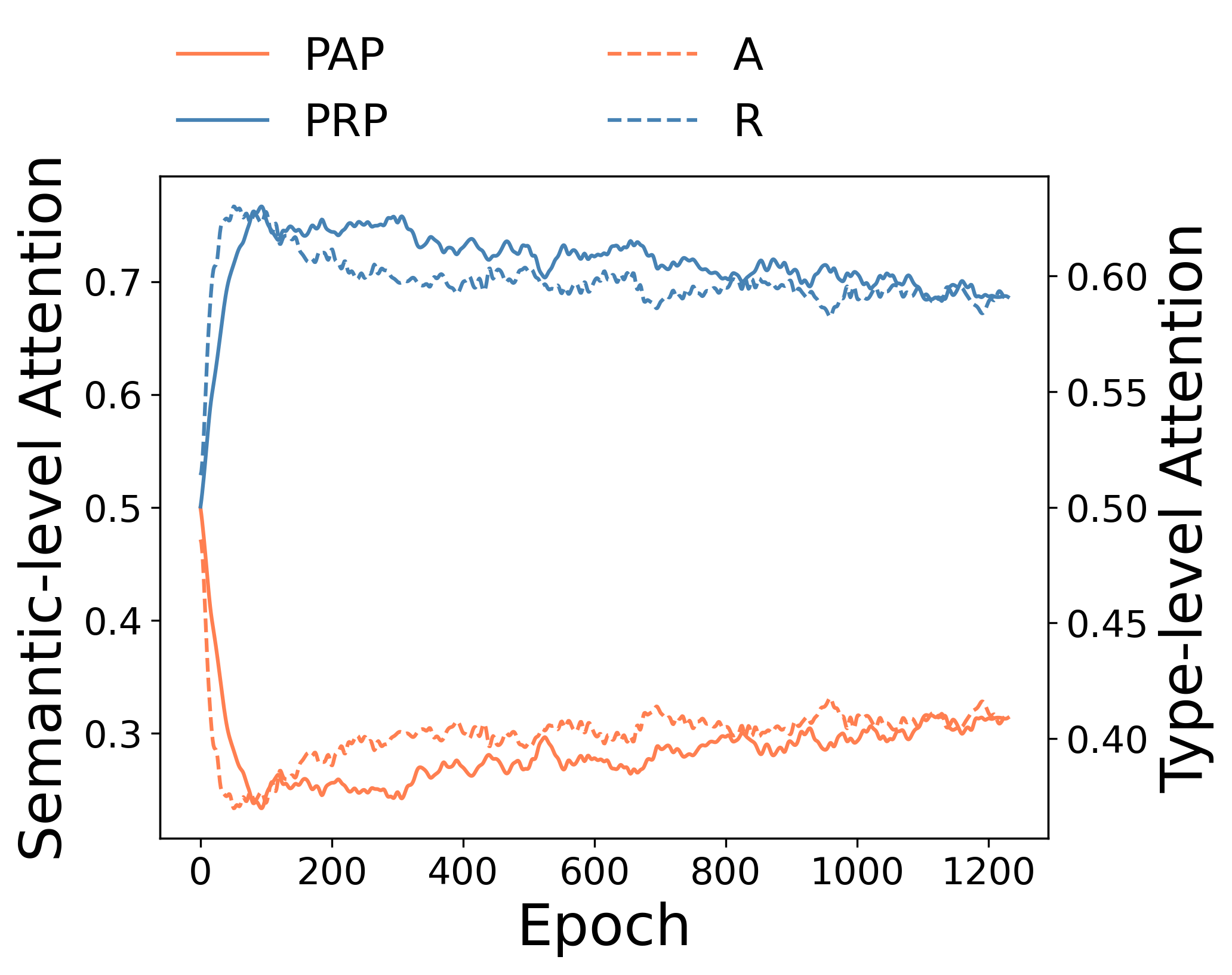}}
\caption{The collaborative changing trends of attentions in two views w.r.t epochs.}
\label{att}
\end{figure}

One salient property of HeCo is the cross-view collaborative mechanism, i.e., HeCo employs the network schema and meta-path views to collaboratively supervise each other to learn the embeddings. In this section, we examine the changing trends of type-level attention $\beta_\Phi$ in network schema view and semantic-level attention $\beta_\mathcal{P}$ in meta-path view w.r.t epochs, and the results are plotted in Figure \ref{att}.
For both ACM and AMiner, the changing trends of two views are collaborative and consistent. Specifically, for ACM, $\beta_\Phi$ of type A is higher than type S, and $\beta_\mathcal{P}$ of meta-path PAP also exceeds that of PSP. For AMiner, type R and meta-path PRP are more important in two views respectively. This indicates that network schema view and meta-path view adapt for each other during training and collaboratively optimize by contrasting each other.

\subsection{Model Extension Analysis}
In this section, we examine results of our extensions. As is shown above, DMGI is a rather competitive method on ACM. So, we compare our two extensions with base model and DMGI on classification and clustering tasks using ACM. The results is shown in Table \ref{kuozhan}.

From the table, we can see that the proposed two versions generally outperform base model and DMGI, especially the version of HeCo\_GAN, which improves the results with a clear margin. As expected, GAN based method can generate harder negatives that are closer to positive distributions. HeCo\_MU is the second best in most cases. The better performance of HeCo\_GAN and HeCo\_MU indicates that more and high-quality negative samples are useful for contrastive learning in general.

\begin{table}[h]
  \caption{Evaluation of extended models on various tasks using ACM (Task 1: Classification; Task 2: Clustering).}
  \label{kuozhan}
  \resizebox{0.47\textwidth}{!}{
  \begin{tabular}{c|c|cccc}
       \hline
       \multicolumn{2}{c|}{Task 1}&DMGI&HeCo&HeCo\_MU&HeCo\_GAN\\
       \hline
       \multirow{3}{*}{Ma}
       &20&87.86$\pm$0.2&88.56$\pm$0.8&88.65$\pm$0.8&\textbf{89.22$\pm$1.1}\\
       &40&86.23$\pm$0.8&87.61$\pm$0.5&87.78$\pm$1.7&\textbf{88.61$\pm$1.6}\\
       &60&87.97$\pm$0.4&89.04$\pm$0.5&\textbf{89.83$\pm$0.5}&89.55$\pm$1.3\\
       \hline
       \multirow{3}{*}{Mi}
       &20&87.60$\pm$0.8&88.13$\pm$0.8&88.39$\pm$0.9&\textbf{88.92$\pm$0.9}\\
       &40&86.02$\pm$0.9&87.45$\pm$0.5&87.66$\pm$1.7&\textbf{88.48$\pm$1.7}\\
       &60&87.82$\pm$0.5&88.71$\pm$0.5&\textbf{89.52$\pm$0.5}&89.29$\pm$1.4\\
       \hline
       \multirow{3}{*}{AUC}
       &20&96.72$\pm$0.3&96.49$\pm$0.3&96.38$\pm$0.5&\textbf{96.91$\pm$0.3}\\
       &40&96.35$\pm$0.3&96.40$\pm$0.4&96.54$\pm$0.5&\textbf{97.13$\pm$0.5}\\
       &60&96.79$\pm$0.2&96.55$\pm$0.3&96.67$\pm$0.7&\textbf{97.12$\pm$0.4}\\
       \hline
       \multicolumn{2}{c|}{Task 2}&DMGI&HeCo&HeCo\_MU&HeCo\_GAN\\
       \hline
       \multicolumn{2}{c|}{NMI}
       &51.66&56.87&58.17&\textbf{59.34}\\
       \multicolumn{2}{c|}{ARI}
       &46.64&56.94&59.41&\textbf{61.48}\\
       \hline
  \end{tabular}}
\end{table}

\subsection{Analysis of Hyper-parameters}
In this section, we systematically investigate the sensitivity of two main parameters: the threshold of positives $T_{pos}$ and the threshold of sampled neighbors with $T_{\Phi_m}$. We conduct node classification on ACM and AMiner datasets and report the Micro-F1 values.

\begin{figure}
\centering
\subfigure[ACM]{
\label{pos_acm}
        \centering
\includegraphics[scale=0.25]{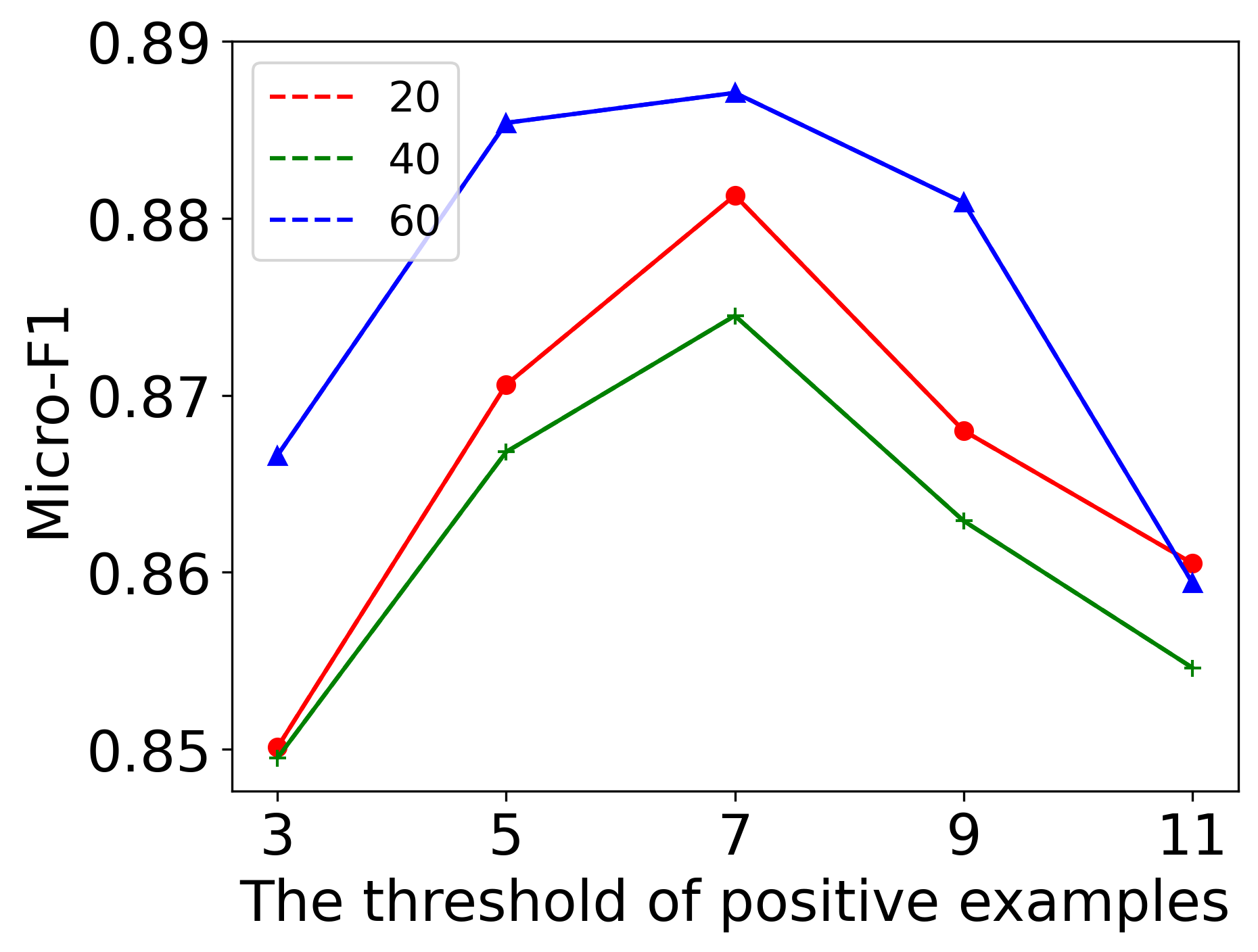}}
\subfigure[AMiner]{
\label{pos_aminer}
        \centering
\includegraphics[scale=0.25]{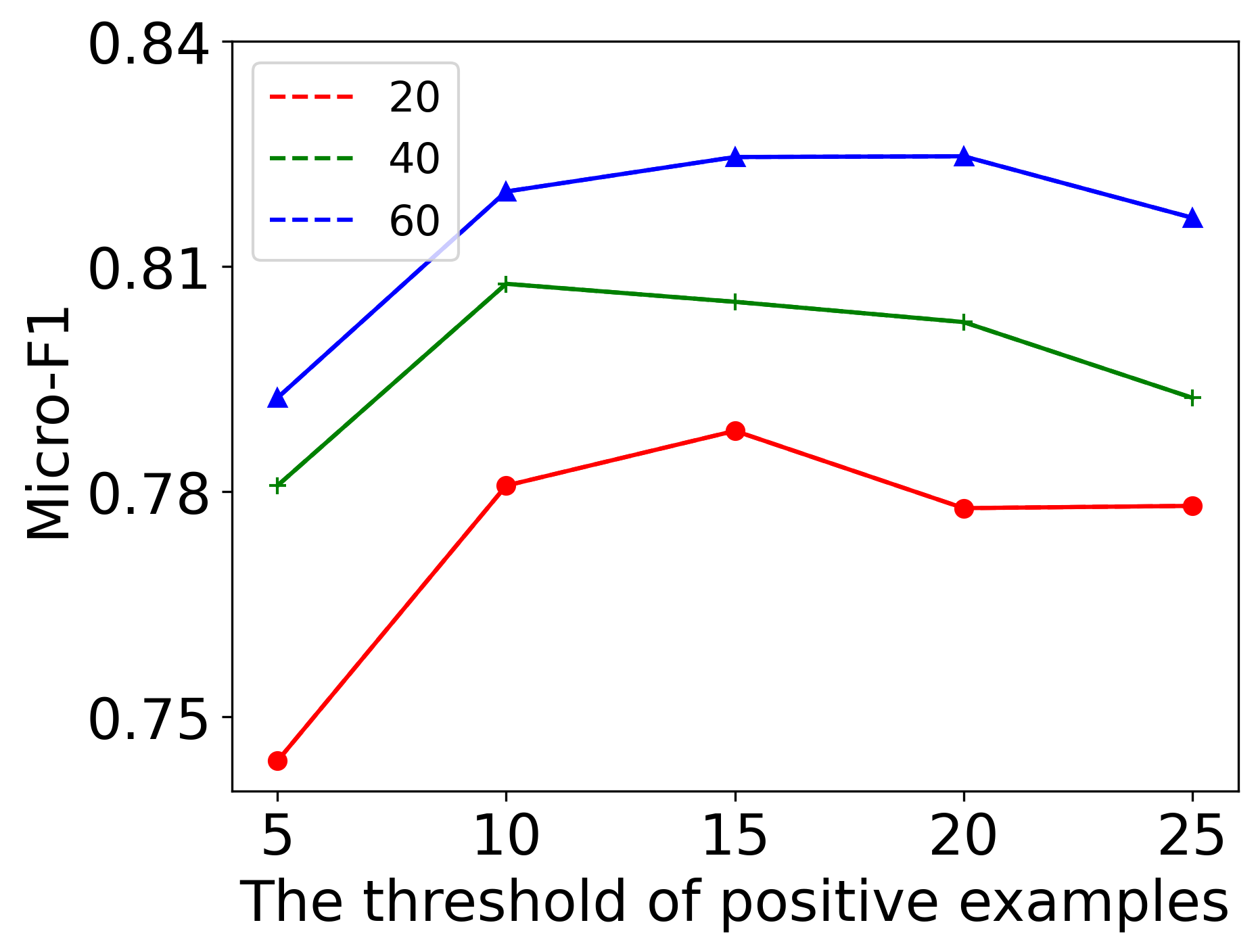}}
\caption{Analysis of the threshold of positive samples.}
\label{pos}
\end{figure}

\textbf{Analysis of $T_{pos}$.} The threshold $T_{pos}$ determines the number of positive samples. We vary the value of it and corresponding results are shown in Figure \ref{pos}. With the increase of $T_{pos}$, the performance goes up first and then declines, and optimum point for ACM is at 7 and at 15 for AMiner. For both datasets, three curves representing different label rates show similar changing trends.

\begin{figure}
\centering
\subfigure[ACM: type A]{
\label{sam_acm}
        \centering
\includegraphics[scale=0.17]{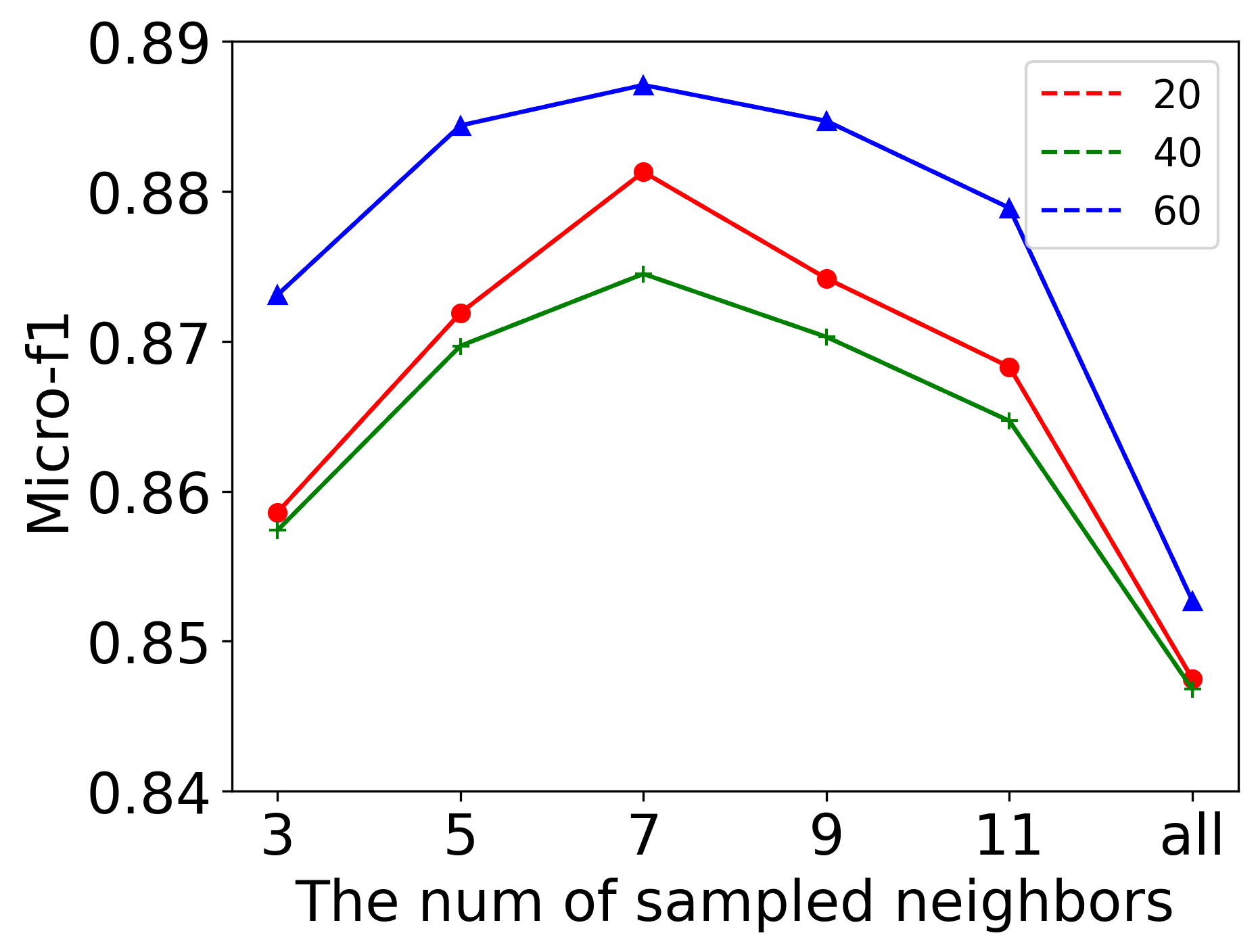}}
\subfigure[AMiner: type A]{
\label{sam_a_aminer}
        \centering
\includegraphics[scale=0.17]{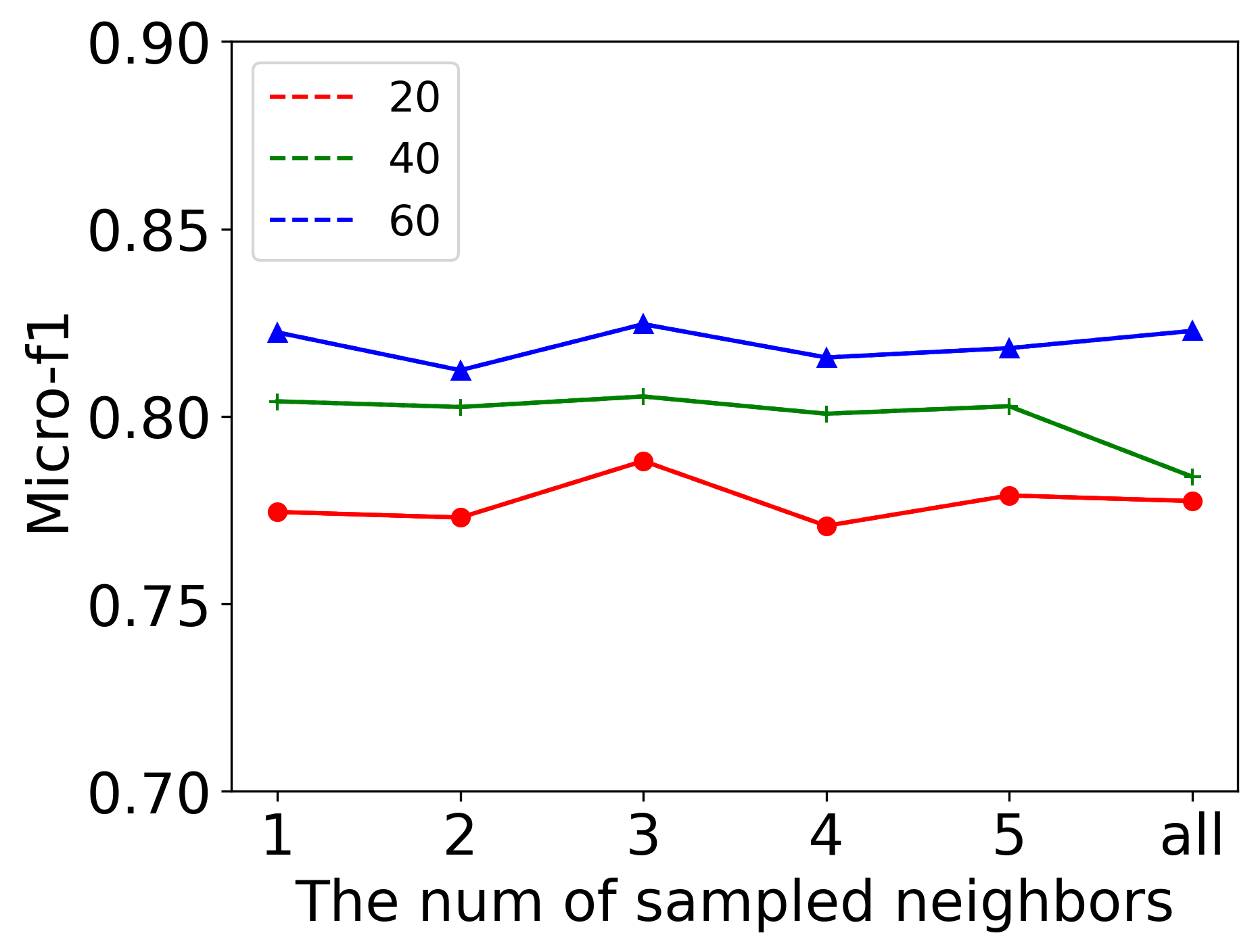}}
\subfigure[AMiner: type R]{
\label{sam_r_aminer}
        \centering
\includegraphics[scale=0.17]{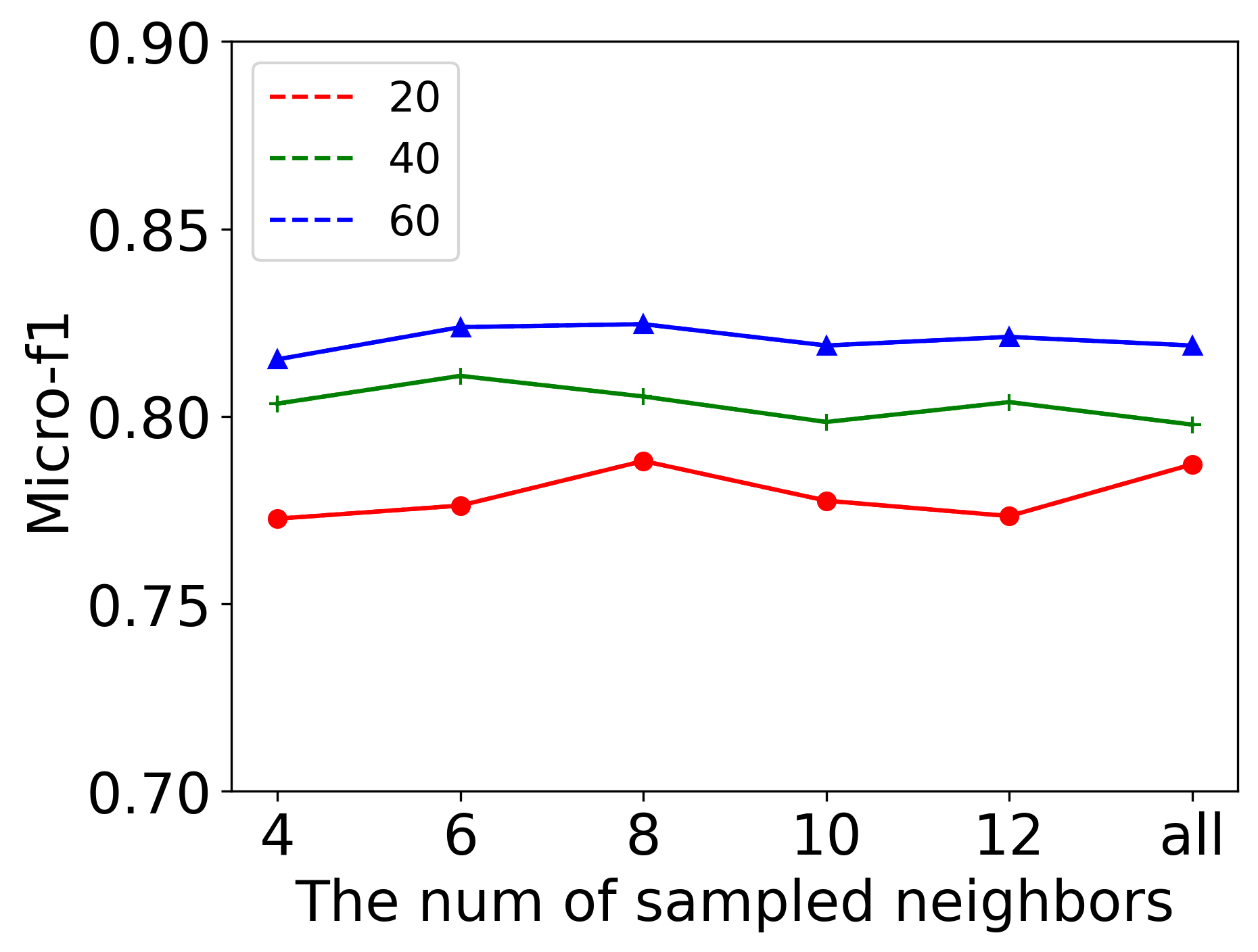}}
\caption{Analysis of the number of sampled neighbors.}
\label{sam}
\end{figure}

\textbf{Analysis of $T_{\Phi_m}$.} To make contrast harder, for target nodes, we randomly sample $T_{\Phi_m}$ neighbors of $\Phi_m$ type, repeatably or not. We again change the value of $T_{\Phi_m}$. It should be pointed out that in ACM, every paper only belongs to one subject (S), so we just change the threshold of type A. The results are shown in Figure \ref{sam}. As can be seen, ACM is sensitive to $T_{\Phi_m}$ of type A, and the best result is achieved when $T_{\Phi_m}=7$. However, AMiner behaves stably with type A or type R. So in our main experiments, we set the $T_{\Phi_m}=3$ for A and $T_{\Phi_m}=8$ for R. Additionally, we also test the case that aggregates all neighbors without sampling, which is marked as "all" in x-axis shown in the figure. In general, "all" cannot perform very well, indicating the usefulness of sampling strategy when we aggregate neighbors.

\section{CONCLUSION}
In this paper, we propose a novel self-supervised heterogeneous graph neural networks with cross-view contrastive learning, named HeCo. HeCo employs network schema and meta-path as two views to capture both of local and high-order structures, and performs the contrastive learning across them. These two views are mutually supervised and finally collaboratively learn the node embeddings. Moreover, a view mask mechanism and two extensions of HeCo are designed to make the contrastive learning harder, so as to further improve the performance of HeCo. Extensive experimental results, as well as the collaboratively changing trends between these two views, verify the effectiveness of HeCo.

\begin{acks}
We thank Zhenyi Wang for the useful discussion. We also thank anonymous reviewers for their time and effort in reviewing this paper. This work is supported in part by the National Natural Science Foundation of China (No. U20B2045, 61772082, 61702296, 62002029). It is also supported by "The Fundamental Research Funds for the Central Universities 2021RC28".
\end{acks}

\bibliographystyle{ACM-Reference-Format}


\begin{thebibliography}{40}


\ifx \showCODEN    \undefined \def \showCODEN     #1{\unskip}     \fi
\ifx \showDOI      \undefined \def \showDOI       #1{#1}\fi
\ifx \showISBNx    \undefined \def \showISBNx     #1{\unskip}     \fi
\ifx \showISBNxiii \undefined \def \showISBNxiii  #1{\unskip}     \fi
\ifx \showISSN     \undefined \def \showISSN      #1{\unskip}     \fi
\ifx \showLCCN     \undefined \def \showLCCN      #1{\unskip}     \fi
\ifx \shownote     \undefined \def \shownote      #1{#1}          \fi
\ifx \showarticletitle \undefined \def \showarticletitle #1{#1}   \fi
\ifx \showURL      \undefined \def \showURL       {\relax}        \fi
\providecommand\bibfield[2]{#2}
\providecommand\bibinfo[2]{#2}
\providecommand\natexlab[1]{#1}
\providecommand\showeprint[2][]{arXiv:#2}

\bibitem[\protect\citeauthoryear{Bachman, Hjelm, and Buchwalter}{Bachman
  et~al\mbox{.}}{2019}]%
        {amdim}
\bibfield{author}{\bibinfo{person}{Philip Bachman}, \bibinfo{person}{R.~Devon
  Hjelm}, {and} \bibinfo{person}{William Buchwalter}.}
  \bibinfo{year}{2019}\natexlab{}.
\newblock \showarticletitle{Learning Representations by Maximizing Mutual
  Information Across Views}. In \bibinfo{booktitle}{\emph{NeurIPS}}.
  \bibinfo{pages}{15509--15519}.
\newblock


\bibitem[\protect\citeauthoryear{Chen, Kornblith, Norouzi, and Hinton}{Chen
  et~al\mbox{.}}{2020}]%
        {simclr}
\bibfield{author}{\bibinfo{person}{Ting Chen}, \bibinfo{person}{Simon
  Kornblith}, \bibinfo{person}{Mohammad Norouzi}, {and}
  \bibinfo{person}{Geoffrey~E. Hinton}.} \bibinfo{year}{2020}\natexlab{}.
\newblock \showarticletitle{A Simple Framework for Contrastive Learning of
  Visual Representations}. In \bibinfo{booktitle}{\emph{ICML}}.
  \bibinfo{pages}{1597--1607}.
\newblock


\bibitem[\protect\citeauthoryear{Davis, Grondin, Johnson, Sciaky, King,
  McMorran, Wiegers, Wiegers, and Mattingly}{Davis et~al\mbox{.}}{2017}]%
        {DBLP:journals/nar/DavisGJSKMWWM17}
\bibfield{author}{\bibinfo{person}{Allan~Peter Davis},
  \bibinfo{person}{Cynthia~J. Grondin}, \bibinfo{person}{Robin~J. Johnson},
  \bibinfo{person}{Daniela Sciaky}, \bibinfo{person}{Benjamin~L. King},
  \bibinfo{person}{Roy McMorran}, \bibinfo{person}{Jolene Wiegers},
  \bibinfo{person}{Thomas~C. Wiegers}, {and} \bibinfo{person}{Carolyn~J.
  Mattingly}.} \bibinfo{year}{2017}\natexlab{}.
\newblock \showarticletitle{The Comparative Toxicogenomics Database: update
  2017}.
\newblock \bibinfo{journal}{\emph{Nucleic Acids Res.}} (\bibinfo{year}{2017}),
  \bibinfo{pages}{D972--D978}.
\newblock


\bibitem[\protect\citeauthoryear{Devlin, Chang, Lee, and Toutanova}{Devlin
  et~al\mbox{.}}{2019}]%
        {nlp2}
\bibfield{author}{\bibinfo{person}{Jacob Devlin}, \bibinfo{person}{Ming{-}Wei
  Chang}, \bibinfo{person}{Kenton Lee}, {and} \bibinfo{person}{Kristina
  Toutanova}.} \bibinfo{year}{2019}\natexlab{}.
\newblock \showarticletitle{{BERT:} Pre-training of Deep Bidirectional
  Transformers for Language Understanding}. In
  \bibinfo{booktitle}{\emph{NAACL-HLT}}. \bibinfo{pages}{4171--4186}.
\newblock


\bibitem[\protect\citeauthoryear{Dong, Chawla, and Swami}{Dong
  et~al\mbox{.}}{2017}]%
        {mp2vec}
\bibfield{author}{\bibinfo{person}{Yuxiao Dong}, \bibinfo{person}{Nitesh~V.
  Chawla}, {and} \bibinfo{person}{Ananthram Swami}.}
  \bibinfo{year}{2017}\natexlab{}.
\newblock \showarticletitle{metapath2vec: Scalable Representation Learning for
  Heterogeneous Networks}. In \bibinfo{booktitle}{\emph{SIGKDD}}.
  \bibinfo{pages}{135--144}.
\newblock


\bibitem[\protect\citeauthoryear{Fan, Zhu, Han, Shi, Hu, Ma, and Li}{Fan
  et~al\mbox{.}}{2019}]%
        {DBLP:conf/kdd/FanZHSHML19}
\bibfield{author}{\bibinfo{person}{Shaohua Fan}, \bibinfo{person}{Junxiong
  Zhu}, \bibinfo{person}{Xiaotian Han}, \bibinfo{person}{Chuan Shi},
  \bibinfo{person}{Linmei Hu}, \bibinfo{person}{Biyu Ma}, {and}
  \bibinfo{person}{Yongliang Li}.} \bibinfo{year}{2019}\natexlab{}.
\newblock \showarticletitle{Metapath-guided Heterogeneous Graph Neural Network
  for Intent Recommendation}. In \bibinfo{booktitle}{\emph{SIGKDD}}.
  \bibinfo{pages}{2478--2486}.
\newblock


\bibitem[\protect\citeauthoryear{Fan, Hou, Zhang, Ye, and Abdulhayoglu}{Fan
  et~al\mbox{.}}{2018}]%
        {DBLP:conf/kdd/FanHZYA18}
\bibfield{author}{\bibinfo{person}{Yujie Fan}, \bibinfo{person}{Shifu Hou},
  \bibinfo{person}{Yiming Zhang}, \bibinfo{person}{Yanfang Ye}, {and}
  \bibinfo{person}{Melih Abdulhayoglu}.} \bibinfo{year}{2018}\natexlab{}.
\newblock \showarticletitle{Gotcha - Sly Malware!: Scorpion {A} Metagraph2vec
  Based Malware Detection System}. In \bibinfo{booktitle}{\emph{SIGKDD}}.
  \bibinfo{pages}{253--262}.
\newblock


\bibitem[\protect\citeauthoryear{Fu, Zhang, Meng, and King}{Fu
  et~al\mbox{.}}{2020}]%
        {magnn}
\bibfield{author}{\bibinfo{person}{Xinyu Fu}, \bibinfo{person}{Jiani Zhang},
  \bibinfo{person}{Ziqiao Meng}, {and} \bibinfo{person}{Irwin King}.}
  \bibinfo{year}{2020}\natexlab{}.
\newblock \showarticletitle{{MAGNN:} Metapath Aggregated Graph Neural Network
  for Heterogeneous Graph Embedding}. In \bibinfo{booktitle}{\emph{WWW}}.
  \bibinfo{pages}{2331--2341}.
\newblock


\bibitem[\protect\citeauthoryear{Glorot and Bengio}{Glorot and Bengio}{2010}]%
        {glorot2010understanding}
\bibfield{author}{\bibinfo{person}{Xavier Glorot} {and} \bibinfo{person}{Yoshua
  Bengio}.} \bibinfo{year}{2010}\natexlab{}.
\newblock \showarticletitle{Understanding the difficulty of training deep
  feedforward neural networks}. In \bibinfo{booktitle}{\emph{AISTATS}}.
  \bibinfo{pages}{249--256}.
\newblock


\bibitem[\protect\citeauthoryear{Goodfellow, Pouget{-}Abadie, Mirza, Xu,
  Warde{-}Farley, Ozair, Courville, and Bengio}{Goodfellow
  et~al\mbox{.}}{2014}]%
        {gan}
\bibfield{author}{\bibinfo{person}{Ian~J. Goodfellow}, \bibinfo{person}{Jean
  Pouget{-}Abadie}, \bibinfo{person}{Mehdi Mirza}, \bibinfo{person}{Bing Xu},
  \bibinfo{person}{David Warde{-}Farley}, \bibinfo{person}{Sherjil Ozair},
  \bibinfo{person}{Aaron~C. Courville}, {and} \bibinfo{person}{Yoshua Bengio}.}
  \bibinfo{year}{2014}\natexlab{}.
\newblock \showarticletitle{Generative adversarial networks}.
\newblock \bibinfo{journal}{\emph{arXiv preprint arXiv:1406.2661}}
  (\bibinfo{year}{2014}).
\newblock


\bibitem[\protect\citeauthoryear{Hamilton, Ying, and Leskovec}{Hamilton
  et~al\mbox{.}}{2017}]%
        {GraphSAGE}
\bibfield{author}{\bibinfo{person}{William~L. Hamilton},
  \bibinfo{person}{Zhitao Ying}, {and} \bibinfo{person}{Jure Leskovec}.}
  \bibinfo{year}{2017}\natexlab{}.
\newblock \showarticletitle{Inductive Representation Learning on Large Graphs}.
  In \bibinfo{booktitle}{\emph{NeurIPS}}. \bibinfo{pages}{1024--1034}.
\newblock


\bibitem[\protect\citeauthoryear{Hassani and Ahmadi}{Hassani and
  Ahmadi}{2020}]%
        {mvgrl}
\bibfield{author}{\bibinfo{person}{Kaveh Hassani} {and} \bibinfo{person}{Amir
  Hosein~Khas Ahmadi}.} \bibinfo{year}{2020}\natexlab{}.
\newblock \showarticletitle{Contrastive Multi-View Representation Learning on
  Graphs}. In \bibinfo{booktitle}{\emph{ICML}}. \bibinfo{pages}{4116--4126}.
\newblock


\bibitem[\protect\citeauthoryear{He, Fan, Wu, Xie, and Girshick}{He
  et~al\mbox{.}}{2020}]%
        {moco}
\bibfield{author}{\bibinfo{person}{Kaiming He}, \bibinfo{person}{Haoqi Fan},
  \bibinfo{person}{Yuxin Wu}, \bibinfo{person}{Saining Xie}, {and}
  \bibinfo{person}{Ross~B. Girshick}.} \bibinfo{year}{2020}\natexlab{}.
\newblock \showarticletitle{Momentum Contrast for Unsupervised Visual
  Representation Learning}. In \bibinfo{booktitle}{\emph{CVPR}}.
  \bibinfo{pages}{9726--9735}.
\newblock


\bibitem[\protect\citeauthoryear{Hu, Fang, and Shi}{Hu et~al\mbox{.}}{2019}]%
        {hegan}
\bibfield{author}{\bibinfo{person}{Binbin Hu}, \bibinfo{person}{Yuan Fang},
  {and} \bibinfo{person}{Chuan Shi}.} \bibinfo{year}{2019}\natexlab{}.
\newblock \showarticletitle{Adversarial Learning on Heterogeneous Information
  Networks}. In \bibinfo{booktitle}{\emph{SIGKDD}}. \bibinfo{pages}{120--129}.
\newblock


\bibitem[\protect\citeauthoryear{Hu, Liu, Gomes, Zitnik, Liang, Pande, and
  Leskovec}{Hu et~al\mbox{.}}{2020b}]%
        {hu2020strategies}
\bibfield{author}{\bibinfo{person}{Weihua Hu}, \bibinfo{person}{Bowen Liu},
  \bibinfo{person}{Joseph Gomes}, \bibinfo{person}{Marinka Zitnik},
  \bibinfo{person}{Percy Liang}, \bibinfo{person}{Vijay~S. Pande}, {and}
  \bibinfo{person}{Jure Leskovec}.} \bibinfo{year}{2020}\natexlab{b}.
\newblock \showarticletitle{Strategies for Pre-training Graph Neural Networks}.
  In \bibinfo{booktitle}{\emph{ICLR}}.
\newblock


\bibitem[\protect\citeauthoryear{Hu, Dong, Wang, and Sun}{Hu
  et~al\mbox{.}}{2020a}]%
        {hgt}
\bibfield{author}{\bibinfo{person}{Ziniu Hu}, \bibinfo{person}{Yuxiao Dong},
  \bibinfo{person}{Kuansan Wang}, {and} \bibinfo{person}{Yizhou Sun}.}
  \bibinfo{year}{2020}\natexlab{a}.
\newblock \showarticletitle{Heterogeneous Graph Transformer}. In
  \bibinfo{booktitle}{\emph{WWW}}. \bibinfo{pages}{2704--2710}.
\newblock


\bibitem[\protect\citeauthoryear{Kalantidis, Sariyildiz, Pion, Weinzaepfel, and
  Larlus}{Kalantidis et~al\mbox{.}}{2020}]%
        {mochi}
\bibfield{author}{\bibinfo{person}{Yannis Kalantidis},
  \bibinfo{person}{Mert~B{\"{u}}lent Sariyildiz}, \bibinfo{person}{No{\'{e}}
  Pion}, \bibinfo{person}{Philippe Weinzaepfel}, {and} \bibinfo{person}{Diane
  Larlus}.} \bibinfo{year}{2020}\natexlab{}.
\newblock \showarticletitle{Hard Negative Mixing for Contrastive Learning}. In
  \bibinfo{booktitle}{\emph{NeurIPS}}.
\newblock


\bibitem[\protect\citeauthoryear{Kingma and Ba}{Kingma and Ba}{2015}]%
        {adam}
\bibfield{author}{\bibinfo{person}{Diederik~P. Kingma} {and}
  \bibinfo{person}{Jimmy Ba}.} \bibinfo{year}{2015}\natexlab{}.
\newblock \showarticletitle{Adam: {A} Method for Stochastic Optimization}. In
  \bibinfo{booktitle}{\emph{ICLR}}.
\newblock


\bibitem[\protect\citeauthoryear{Kipf and Welling}{Kipf and Welling}{2016}]%
        {gae}
\bibfield{author}{\bibinfo{person}{Thomas~N. Kipf} {and} \bibinfo{person}{Max
  Welling}.} \bibinfo{year}{2016}\natexlab{}.
\newblock \showarticletitle{Variational graph auto-encoders}.
\newblock \bibinfo{journal}{\emph{arXiv preprint arXiv:1611.07308}}
  (\bibinfo{year}{2016}).
\newblock


\bibitem[\protect\citeauthoryear{Kipf and Welling}{Kipf and Welling}{2017}]%
        {gcn}
\bibfield{author}{\bibinfo{person}{Thomas~N. Kipf} {and} \bibinfo{person}{Max
  Welling}.} \bibinfo{year}{2017}\natexlab{}.
\newblock \showarticletitle{Semi-Supervised Classification with Graph
  Convolutional Networks}. In \bibinfo{booktitle}{\emph{ICLR}}.
\newblock


\bibitem[\protect\citeauthoryear{Lan, Chen, Goodman, Gimpel, Sharma, and
  Soricut}{Lan et~al\mbox{.}}{2020}]%
        {nlp1}
\bibfield{author}{\bibinfo{person}{Zhenzhong Lan}, \bibinfo{person}{Mingda
  Chen}, \bibinfo{person}{Sebastian Goodman}, \bibinfo{person}{Kevin Gimpel},
  \bibinfo{person}{Piyush Sharma}, {and} \bibinfo{person}{Radu Soricut}.}
  \bibinfo{year}{2020}\natexlab{}.
\newblock \showarticletitle{{ALBERT:} {A} Lite {BERT} for Self-supervised
  Learning of Language Representations}. In \bibinfo{booktitle}{\emph{ICLR}}.
\newblock


\bibitem[\protect\citeauthoryear{Li, Ding, Kao, Sun, and Mamoulis}{Li
  et~al\mbox{.}}{2020}]%
        {freebase}
\bibfield{author}{\bibinfo{person}{Xiang Li}, \bibinfo{person}{Danhao Ding},
  \bibinfo{person}{Ben Kao}, \bibinfo{person}{Yizhou Sun}, {and}
  \bibinfo{person}{Nikos Mamoulis}.} \bibinfo{year}{2020}\natexlab{}.
\newblock \showarticletitle{Leveraging Meta-path Contexts for Classification in
  Heterogeneous Information Networks}.
\newblock \bibinfo{journal}{\emph{arXiv preprint arXiv:2012.10024}}
  (\bibinfo{year}{2020}).
\newblock


\bibitem[\protect\citeauthoryear{Linsker}{Linsker}{1988}]%
        {infomax}
\bibfield{author}{\bibinfo{person}{Ralph Linsker}.}
  \bibinfo{year}{1988}\natexlab{}.
\newblock \showarticletitle{Self-Organization in a Perceptual Network}.
\newblock \bibinfo{journal}{\emph{Computer}} (\bibinfo{year}{1988}),
  \bibinfo{pages}{105--117}.
\newblock


\bibitem[\protect\citeauthoryear{Liu, Zhang, Hou, Wang, Mian, Zhang, and
  Tang}{Liu et~al\mbox{.}}{2020}]%
        {liu2020self}
\bibfield{author}{\bibinfo{person}{Xiao Liu}, \bibinfo{person}{Fanjin Zhang},
  \bibinfo{person}{Zhenyu Hou}, \bibinfo{person}{Zhaoyu Wang},
  \bibinfo{person}{Li Mian}, \bibinfo{person}{Jing Zhang}, {and}
  \bibinfo{person}{Jie Tang}.} \bibinfo{year}{2020}\natexlab{}.
\newblock \showarticletitle{Self-supervised learning: Generative or
  contrastive}.
\newblock \bibinfo{journal}{\emph{arXiv preprint arXiv:2006.08218}}
  (\bibinfo{year}{2020}).
\newblock


\bibitem[\protect\citeauthoryear{Oord, Li, and Vinyals}{Oord
  et~al\mbox{.}}{2018}]%
        {cpc}
\bibfield{author}{\bibinfo{person}{Aaron van~den Oord}, \bibinfo{person}{Yazhe
  Li}, {and} \bibinfo{person}{Oriol Vinyals}.} \bibinfo{year}{2018}\natexlab{}.
\newblock \showarticletitle{Representation learning with contrastive predictive
  coding}.
\newblock \bibinfo{journal}{\emph{arXiv preprint arXiv:1807.03748}}
  (\bibinfo{year}{2018}).
\newblock


\bibitem[\protect\citeauthoryear{Park, Kim, Han, and Yu}{Park
  et~al\mbox{.}}{2020}]%
        {dmgi}
\bibfield{author}{\bibinfo{person}{Chanyoung Park}, \bibinfo{person}{Donghyun
  Kim}, \bibinfo{person}{Jiawei Han}, {and} \bibinfo{person}{Hwanjo Yu}.}
  \bibinfo{year}{2020}\natexlab{}.
\newblock \showarticletitle{Unsupervised Attributed Multiplex Network
  Embedding}. In \bibinfo{booktitle}{\emph{AAAI}}. \bibinfo{pages}{5371--5378}.
\newblock


\bibitem[\protect\citeauthoryear{Peng, Huang, Luo, Zheng, Rong, Xu, and
  Huang}{Peng et~al\mbox{.}}{2020}]%
        {gmi}
\bibfield{author}{\bibinfo{person}{Zhen Peng}, \bibinfo{person}{Wenbing Huang},
  \bibinfo{person}{Minnan Luo}, \bibinfo{person}{Qinghua Zheng},
  \bibinfo{person}{Yu Rong}, \bibinfo{person}{Tingyang Xu}, {and}
  \bibinfo{person}{Junzhou Huang}.} \bibinfo{year}{2020}\natexlab{}.
\newblock \showarticletitle{Graph Representation Learning via Graphical Mutual
  Information Maximization}. In \bibinfo{booktitle}{\emph{WWW}}.
  \bibinfo{pages}{259--270}.
\newblock


\bibitem[\protect\citeauthoryear{Qiu, Chen, Dong, Zhang, Yang, Ding, Wang, and
  Tang}{Qiu et~al\mbox{.}}{2020}]%
        {gcc}
\bibfield{author}{\bibinfo{person}{Jiezhong Qiu}, \bibinfo{person}{Qibin Chen},
  \bibinfo{person}{Yuxiao Dong}, \bibinfo{person}{Jing Zhang},
  \bibinfo{person}{Hongxia Yang}, \bibinfo{person}{Ming Ding},
  \bibinfo{person}{Kuansan Wang}, {and} \bibinfo{person}{Jie Tang}.}
  \bibinfo{year}{2020}\natexlab{}.
\newblock \showarticletitle{{GCC:} Graph Contrastive Coding for Graph Neural
  Network Pre-Training}. In \bibinfo{booktitle}{\emph{KDD}}.
  \bibinfo{pages}{1150--1160}.
\newblock


\bibitem[\protect\citeauthoryear{Shi, Hu, Zhao, and Yu}{Shi
  et~al\mbox{.}}{2019}]%
        {herec}
\bibfield{author}{\bibinfo{person}{Chuan Shi}, \bibinfo{person}{Binbin Hu},
  \bibinfo{person}{Wayne~Xin Zhao}, {and} \bibinfo{person}{Philip~S. Yu}.}
  \bibinfo{year}{2019}\natexlab{}.
\newblock \showarticletitle{Heterogeneous Information Network Embedding for
  Recommendation}.
\newblock \bibinfo{journal}{\emph{{IEEE} Trans. Knowl. Data Eng.}}
  (\bibinfo{year}{2019}), \bibinfo{pages}{357--370}.
\newblock


\bibitem[\protect\citeauthoryear{Sun and Han}{Sun and Han}{2012}]%
        {DBLP:journals/sigkdd/SunH12}
\bibfield{author}{\bibinfo{person}{Yizhou Sun} {and} \bibinfo{person}{Jiawei
  Han}.} \bibinfo{year}{2012}\natexlab{}.
\newblock \showarticletitle{Mining heterogeneous information networks: a
  structural analysis approach}.
\newblock \bibinfo{journal}{\emph{{SIGKDD} Explor.}} (\bibinfo{year}{2012}),
  \bibinfo{pages}{20--28}.
\newblock


\bibitem[\protect\citeauthoryear{Sun, Han, Yan, Yu, and Wu}{Sun
  et~al\mbox{.}}{2011}]%
        {pathsim}
\bibfield{author}{\bibinfo{person}{Yizhou Sun}, \bibinfo{person}{Jiawei Han},
  \bibinfo{person}{Xifeng Yan}, \bibinfo{person}{Philip~S. Yu}, {and}
  \bibinfo{person}{Tianyi Wu}.} \bibinfo{year}{2011}\natexlab{}.
\newblock \showarticletitle{PathSim: Meta Path-Based Top-K Similarity Search in
  Heterogeneous Information Networks}.
\newblock \bibinfo{journal}{\emph{VLDB.}} (\bibinfo{year}{2011}),
  \bibinfo{pages}{992--1003}.
\newblock


\bibitem[\protect\citeauthoryear{Tian, Krishnan, and Isola}{Tian
  et~al\mbox{.}}{2020}]%
        {cmc}
\bibfield{author}{\bibinfo{person}{Yonglong Tian}, \bibinfo{person}{Dilip
  Krishnan}, {and} \bibinfo{person}{Phillip Isola}.}
  \bibinfo{year}{2020}\natexlab{}.
\newblock \showarticletitle{Contrastive Multiview Coding}. In
  \bibinfo{booktitle}{\emph{ECCV}}. \bibinfo{pages}{776--794}.
\newblock


\bibitem[\protect\citeauthoryear{Velickovic, Fedus, Hamilton, Li{\`{o}},
  Bengio, and Hjelm}{Velickovic et~al\mbox{.}}{2019}]%
        {dgi}
\bibfield{author}{\bibinfo{person}{Petar Velickovic}, \bibinfo{person}{William
  Fedus}, \bibinfo{person}{William~L. Hamilton}, \bibinfo{person}{Pietro
  Li{\`{o}}}, \bibinfo{person}{Yoshua Bengio}, {and} \bibinfo{person}{R.~Devon
  Hjelm}.} \bibinfo{year}{2019}\natexlab{}.
\newblock \showarticletitle{Deep Graph Infomax}. In
  \bibinfo{booktitle}{\emph{ICLR}}.
\newblock


\bibitem[\protect\citeauthoryear{Wang, Wang, Wang, Zhao, Zhang, Zhang, Xie, and
  Guo}{Wang et~al\mbox{.}}{2018}]%
        {graphgan}
\bibfield{author}{\bibinfo{person}{Hongwei Wang}, \bibinfo{person}{Jia Wang},
  \bibinfo{person}{Jialin Wang}, \bibinfo{person}{Miao Zhao},
  \bibinfo{person}{Weinan Zhang}, \bibinfo{person}{Fuzheng Zhang},
  \bibinfo{person}{Xing Xie}, {and} \bibinfo{person}{Minyi Guo}.}
  \bibinfo{year}{2018}\natexlab{}.
\newblock \showarticletitle{GraphGAN: Graph Representation Learning With
  Generative Adversarial Nets}. In \bibinfo{booktitle}{\emph{AAAI}}.
  \bibinfo{pages}{2508--2515}.
\newblock


\bibitem[\protect\citeauthoryear{Wang, Ji, Shi, Wang, Ye, Cui, and Yu}{Wang
  et~al\mbox{.}}{2019}]%
        {han}
\bibfield{author}{\bibinfo{person}{Xiao Wang}, \bibinfo{person}{Houye Ji},
  \bibinfo{person}{Chuan Shi}, \bibinfo{person}{Bai Wang},
  \bibinfo{person}{Yanfang Ye}, \bibinfo{person}{Peng Cui}, {and}
  \bibinfo{person}{Philip~S. Yu}.} \bibinfo{year}{2019}\natexlab{}.
\newblock \showarticletitle{Heterogeneous Graph Attention Network}. In
  \bibinfo{booktitle}{\emph{WWW}}. \bibinfo{pages}{2022--2032}.
\newblock


\bibitem[\protect\citeauthoryear{Wu, Pan, Chen, Long, Zhang, and Yu}{Wu
  et~al\mbox{.}}{2021}]%
        {wu2021comprehensive}
\bibfield{author}{\bibinfo{person}{Zonghan Wu}, \bibinfo{person}{Shirui Pan},
  \bibinfo{person}{Fengwen Chen}, \bibinfo{person}{Guodong Long},
  \bibinfo{person}{Chengqi Zhang}, {and} \bibinfo{person}{Philip~S. Yu}.}
  \bibinfo{year}{2021}\natexlab{}.
\newblock \showarticletitle{A Comprehensive Survey on Graph Neural Networks}.
\newblock \bibinfo{journal}{\emph{{IEEE} Trans. Neural Networks Learn. Syst.}}
  (\bibinfo{year}{2021}), \bibinfo{pages}{4--24}.
\newblock


\bibitem[\protect\citeauthoryear{Yun, Jeong, Kim, Kang, and Kim}{Yun
  et~al\mbox{.}}{2019}]%
        {gtn}
\bibfield{author}{\bibinfo{person}{Seongjun Yun}, \bibinfo{person}{Minbyul
  Jeong}, \bibinfo{person}{Raehyun Kim}, \bibinfo{person}{Jaewoo Kang}, {and}
  \bibinfo{person}{Hyunwoo~J. Kim}.} \bibinfo{year}{2019}\natexlab{}.
\newblock \showarticletitle{Graph Transformer Networks}. In
  \bibinfo{booktitle}{\emph{NeurIPS}}. \bibinfo{pages}{11960--11970}.
\newblock


\bibitem[\protect\citeauthoryear{Zhang, Song, Huang, Swami, and Chawla}{Zhang
  et~al\mbox{.}}{2019}]%
        {hetegnn}
\bibfield{author}{\bibinfo{person}{Chuxu Zhang}, \bibinfo{person}{Dongjin
  Song}, \bibinfo{person}{Chao Huang}, \bibinfo{person}{Ananthram Swami}, {and}
  \bibinfo{person}{Nitesh~V. Chawla}.} \bibinfo{year}{2019}\natexlab{}.
\newblock \showarticletitle{Heterogeneous Graph Neural Network}. In
  \bibinfo{booktitle}{\emph{SIGKDD}}. \bibinfo{pages}{793--803}.
\newblock


\bibitem[\protect\citeauthoryear{Zhang, Ciss{\'{e}}, Dauphin, and
  Lopez{-}Paz}{Zhang et~al\mbox{.}}{2018}]%
        {mixup}
\bibfield{author}{\bibinfo{person}{Hongyi Zhang}, \bibinfo{person}{Moustapha
  Ciss{\'{e}}}, \bibinfo{person}{Yann~N. Dauphin}, {and} \bibinfo{person}{David
  Lopez{-}Paz}.} \bibinfo{year}{2018}\natexlab{}.
\newblock \showarticletitle{mixup: Beyond Empirical Risk Minimization}. In
  \bibinfo{booktitle}{\emph{ICLR}}.
\newblock


\bibitem[\protect\citeauthoryear{Zhao, Wang, Shi, Liu, and Ye}{Zhao
  et~al\mbox{.}}{2020}]%
        {nshe}
\bibfield{author}{\bibinfo{person}{Jianan Zhao}, \bibinfo{person}{Xiao Wang},
  \bibinfo{person}{Chuan Shi}, \bibinfo{person}{Zekuan Liu}, {and}
  \bibinfo{person}{Yanfang Ye}.} \bibinfo{year}{2020}\natexlab{}.
\newblock \showarticletitle{Network Schema Preserving Heterogeneous Information
  Network Embedding}. In \bibinfo{booktitle}{\emph{IJCAI}}.
  \bibinfo{pages}{1366--1372}.
\newblock


\end{thebibliography}

\newpage
\appendix

\section{SUPPLEMENT}
In the supplement, for the reproducibility, we provide all the baselines and datasets websites. The implementation details, including the detailed hyper-parameter values, are also provided.
\subsection{Baselines}
The publicly available implementations of baselines can be found at the following URLs:
\begin{itemize}
    \item GraphSAGE: \url{https://github.com/williamleif/GraphSAGE}
    \item GAE: \url{https://github.com/tkipf/gae}
    \item Mp2vec: \url{https://ericdongyx.github.io/metapath2vec/m2v.html}
    \item HERec: \url{https://github.com/librahu/HERec}
    \item HetGNN: \url{https://github.com/chuxuzhang/KDD2019_HetGNN}
    \item HAN: \url{https://github.com/Jhy1993/HAN}
    \item DGI: \url{https://github.com/PetarV-/DGI}
    \item DMGI: \url{https://github.com/pcy1302/DMGI}
\end{itemize}
\subsection{Datasets}
The datasets used in experiments can be found in these URLs:
\begin{itemize}
    \item ACM: \url{https://github.com/Andy-Border/NSHE}
    \item DBLP: \url{https://github.com/cynricfu/MAGNN}
    \item Freebase: \url{https://github.com/dingdanhao110/Conch}
    \item AMiner: \url{https://github.com/librahu/HIN-Datasets-for-Recommen-dation-and-Network-Embedding}
\end{itemize}
\begin{table*}[t]
  \caption{The values of parameter used in HeCo.}
  \label{para}
  \begin{tabular}{c||ccccccc}
       \hline
       Dataset & \textit{lr} & \textit{patience} & \textit{sample\_num} & $\tau$ & \textit{dropout\_feat} & \textit{dropout\_attn} & \textit{weight\_decay}\\
       \hline
       ACM & 0.0008 & 5 & A:7\ ;\ S:1 & 0.8 & 0.3 & 0.5 & 0.0\\
       \hline
       DBLP & 0.0008 & 30 & P:6 & 0.9 & 0.4 & 0.35 & 0.0 \\
       \hline
       Freebase & 0.001 & 20 & D:1\ ;\ A:18\ ;\ W:2 & 0.5 & 0.1 & 0.3 & 0.0 \\
       \hline
       AMiner & 0.003 & 40 & A:3\ ;\ R:8 & 0.5 & 0.5 & 0.5 & 0.0\\
       \hline
  \end{tabular}
\end{table*}
\subsection{Implementation Details}
\label{detail}
We implement HeCo in PyTorch, and list some important parameter values used in our model in Table \ref{para}. In this table, \textit{lr} is the learning rate, and \textit{sample\_num} is $T_{\Phi_m}$, the threshold of sampled neighbors of type $\Phi_m$. It should be pointed that author only connects with paper in the network schema of DBLP, so we only set the threshold for type P. \textit{dropout\_feat} is the dropout value used on projected features, and \textit{dropout\_attn} is the dropout of attentions in two views.

\section{details of H\lowercase{e}C\lowercase{o}\_GAN}
\label{hecogan}
In this section, we further explain the training process of HeCo\_GAN, proposed in section \ref{extension}.

As mentioned above, HeCo\_GAN contains the proposed HeCo, a discriminator D and a generator G. At the beginning of the train, the parameters of HeCo should be warmed up to improve the quality of generated embeddings. So, we first only train HeCo for $K_0$ epochs, which is a hyper-parameter. Then, we get $z^{sc}$ and $z^{mp}$ and utilize them to train D and G alternatively, which is as following two steps:

\begin{itemize}
    \item Freeze G and train D for $K_D$ epochs. For target node i and its embedding $z_i^{sc}$ under network schema view, we can get the embeddings of nodes in $\mathbb{P}_i$ under meta-path view. D outputs a probability that a sample $j$ is from $\mathbb{P}_i$ given $z_i^{sc}$:
    \begin{equation}
        D\left(z_j|z_i^{sc}\right)=\frac{1}{1+\exp\left(-{z_i^{sc}}^\top M^D_{mp}z_j\right)},
    \end{equation}
    where $M^D_{mp}$ is a matrix that projects $z_i^{sc}$ into the space of meta-path view. And the objective function of D under the network schema view is:
    \begin{equation}
    \begin{aligned}
        \mathcal{L}_{i_D}^{sc}=&-\mathop{\mathbb{E}}\limits_{j\sim\mathbbm{p}_i}\log D\left(z_j^{mp}|z_i^{sc}\right)\\
        &-\mathop{\mathbb{E}}\limits_{\widetilde{z_i^{mp}}\sim G\left(z_i^{sc}\right)}\log\left(1-D\left(\widetilde{z_i^{mp}}|z_i^{sc}\right)\right),
    \end{aligned}
    \end{equation}
    where $\mathbbm{p}_i\subset\mathbb{P}_i$, which is chosen randomly, and $\widetilde{z_i^{mp}}$ is generated by generator based on $z_i^{sc}$. This shows that given $z_i^{sc}$, D aims to identify its positive samples from meta-path view as positive and samples generated by G as negative. Notice that the number of fake samples from G is the same as $|\mathbbm{p}_i|$. Similarly, we can also get the objective function of D under the meta-path view $\mathcal{L}_{i_D}^{mp}$. So, we train the discriminator D by minimizing the following loss:
    \begin{equation}
        \mathcal{L}_{D}=\frac{1}{|B|}\sum\limits_{i\in B}\frac{1}{2}\left(\mathcal{L}_{i_D}^{sc}+\mathcal{L}_{i_D}^{mp}\right),
    \end{equation}
    where $B$ denotes the batch of nodes that are trained in current epoch.

    \item Freeze D and train G for $K_G$ epochs. G gradually improves the quality of generated samples by fooling D. Specifically, given the target $i$ and its embedding $z_i^{sc}$ under network schema view, G first constructs a Gaussian distribution center on $i$, and draws samples from it, which is related to $z_i^{sc}$:
    \begin{equation}
        e^{mp}_{i}\sim\mathcal{N}\left({z_i^{sc}}^\top M^G_{mp}, \sigma^2\textbf{I}\right),
    \end{equation}
    where $M^G_{mp}$ is also a projected function to map $z_i^{sc}$ into meta-path space, and $\sigma^2\textbf{I}$ is covariance. We then apply one-layer MLP to enhance the expression of the fake samples:
    \begin{equation}
        \widetilde{z^{mp}_{i}}=G\left(z_i^{sc}\right)=\sigma\left(We^{mp}_{i}+b\right).
    \end{equation}
    Here, $\sigma$, $W$ and $b$ denote non-linear activation, weight matrix and bias vector, respectively. To fool the discriminator, generator is trained under network schema view by following loss:
    \begin{equation}
    \begin{aligned}
        \mathcal{L}_{i_G}^{sc}&=-\mathop{\mathbb{E}}\limits_{\widetilde{z^{mp}_{i}}\sim G\left(z_i^{sc}\right)}\log D(\widetilde{z^{mp}_{i}}|z_i^{sc}),\\
        \mathcal{L}_{G}&=\frac{1}{|B|}\sum\limits_{i\sim B}\frac{1}{2}\left(\mathcal{L}_{i_G}^{sc}+\mathcal{L}_{i_G}^{mp}\right).
    \end{aligned}
    \end{equation}
    Again, $\mathcal{L}_{i_G}^{mp}$ is attained like $\mathcal{L}_{i_G}^{sc}$.
\end{itemize}

\noindent These two steps are alternated for $I_{DG}$ times to fully train the D and G.

Once we get the well-trained G, high-quality negative samples $\widetilde{z^{mp}_i}$ and $\widetilde{z^{sc}_i}$ will be obtained, given $z^{sc}_i$ and $z^{mp}_i$, respectively. And they are combined with original negative samples from meta-path view or network schema view. Finally, the extended set of negative samples is fed into HeCo to boost the training for $K_{H}$ epochs.

The training processes of the proposed HeCo, discriminator D and generator G are employed iteratively until to the convergence.

\end{document}